\documentclass[10pt,twocolumn,letterpaper]{article}

\usepackage[pagenumbers]{cvpr} 
\usepackage{wrapfig}
\usepackage{multirow}
\usepackage{algorithm}
\usepackage{algorithmic}
\usepackage{amsmath}
\usepackage{mathtools}
\usepackage{amssymb}
\usepackage{amsthm}

\newtheorem{definition}{Definition}
\newtheorem{proposition}{Proposition}

\usepackage[dvipsnames]{xcolor}

\definecolor{cvprblue}{rgb}{0.21,0.49,0.74}
\usepackage[pagebackref,breaklinks,colorlinks,citecolor=cvprblue]{hyperref}

\def\paperID{823} 
\def\confName{CVPR}
\def\confYear{2024}

\title{TPFL: A Trustworthy Personalized Federated Learning Framework via Subjective Logic}

\author{Jinqian Chen\\
Xi'an Jiaotong University\\
Xi'an, China\\
{\tt\small chenjinqian@stu.xjtu.edu.cn}
\and
Jihua Zhu\\
Xi'an Jiaotong University\\
Xi'an, China\\
{\tt\small zhujh@xjtu.edu.cn}
}

\begin{document}
\maketitle
\begin{abstract}
Federated learning (FL) enables collaborative model training across distributed clients while preserving data privacy. Despite its widespread adoption, most FL approaches focusing solely on privacy protection fall short in scenarios where trustworthiness is crucial, necessitating advancements in secure training, dependable decision-making mechanisms, robustness on corruptions, and enhanced performance with Non-IID data. To bridge this gap, we introduce Trustworthy Personalized Federated Learning (TPFL) framework designed for classification tasks via subjective logic in this paper. Specifically, TPFL adopts a unique approach by employing subjective logic to construct federated models, providing probabilistic decisions coupled with an assessment of uncertainty rather than mere probability assignments. By incorporating a trainable heterogeneity prior to the local training phase, TPFL effectively mitigates the adverse effects of data heterogeneity. Model uncertainty and instance uncertainty are further utilized to ensure the safety and reliability of the training and inference stages. Through extensive experiments on widely recognized federated learning benchmarks, we demonstrate that TPFL not only achieves competitive performance compared with advanced methods but also exhibits resilience against prevalent malicious attacks, robustness on domain shifts, and reliability in high-stake scenarios.

\end{abstract}    

\section{Introduction}

The advent of artificial intelligence (AI) has led to the widespread deployment of deep learning models across a myriad of domains \citep{drive, medical, recommend}. Despite this progress, the integration of AI into various fields has unveiled significant challenges, including concerns over data privacy \citep{dataprivacy}, the black-box nature of decision-making \citep{blackbox1, blackbox2}, and vulnerability to adversarial attacks \citep{adver1, adver2}. These issues compromise the potential of AI to autonomously perform tasks traditionally requiring human intervention, underscoring the urgent need for developing trustworthy AI systems. Trustworthy AI represents an ideal paradigm, aiming to harness the benefits of AI technologies while eliminating risks and ensuring safety, fairness, privacy, explainability, and robustness \citep{trustworthyAIsurvey1, trustworthyAISurvey2}. Among the initiatives aimed at realizing trustworthy AI, federated learning (FL) has emerged as a critical branch of focus \citep{FLSurvey}. By enabling decentralized model training across distributed data sources without centralizing data, FL addresses privacy concerns inherent in traditional model training \citep{FedAvg}. This approach has found successful applications in many data-sensitive scenarios such as healthcare \citep{FLMedical}, finance \citep{FLFinance}, and city management \citep{FLCity}.

\begin{figure}
    \centering
    \includegraphics[width=0.47\textwidth]{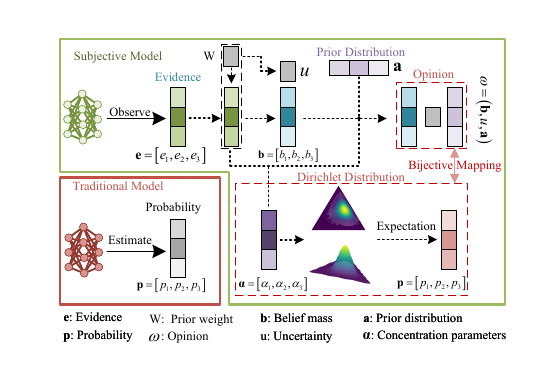}
    \caption{Difference Between Subjective Models and Traditional Models. Traditional models directly estimate the first-order probability distribution through Softmax operation. In contrast, subjective models observe evidence from the instance to form opinions, constructing a Dirichlet distribution (i.e., second-order distribution) on class assignments to allow the estimation of uncertainty. }
    \label{fig1}
\end{figure}

Despite its potential, FL is confronted with significant challenges in real-world applications, notably data heterogeneity \cite{FedProx, FLSurvey2} and susceptibility to adversarial attacks \citep{TFLSurvey, AttackTail}. Recent studies have shown heterogeneity in client data distributions in practical scenarios, including label and feature distribution skews \citep{FLNonIIDSurvey}, can significantly hinder the training of FL, leading to degraded model performance and extended convergence time \citep{ExperimentalStudy}. Additionally, FL's open, collaborative nature makes it susceptible to adversarial attacks by malicious participants \citep{AttackTail, FLThreatSurvey}. These attackers can engage in collaboration with little to no scrutiny, threatening the system at any stage—from data pre-processing and collaborative training to model deployment \citep{TFLSurvey}. They can introduce backdoors into the training models, causing specific instances to be incorrectly classified into a targeted class \citep{BackdoorFL, TargetedBackdoor}, or broadly impair the model's performance across all classes  \citep{FlipAttack, BackdoorFL}. Such attacks, executed through data \citep{ModelReplacement, FlipAttack} or model poisoning strategies \citep{LocalModelPoisoning}, severely hamper the integrity and efficacy of federated learning systems. {To mitigate their malicious effect, many defense methods have emerged based on various ideas, such as robust aggregation \citep{TrimmedMean, Krum}, anomaly detection \citep{AUROR, ClusterFL}, and modification on learning ratios \citep{RobustLR}, etc.} Though significant progress has achieved, current defense methods always involve the trade-off between accuracy and security \citep{DPHarm, ThreatSurvey}, making the performance even worse under Non-IID data.

Moreover, utilizing machine learning models solely is always high risk during the decision-making procedures. An intuitive solution is to manually intervene only when the models are uncertain about their decisions or unable to make decisions. However, such a scenario is just wishful thinking. The fact is that data-driven models always fail to assign proper uncertainty to their predictions, exhibiting severe overconfidence \citep{ECE, Dropout}. Even for an out-of-distribution (OOD) instance, models will confidently assign it to a specific class, lacking the ability to know what they do not know \citep{EDL, UncertaintySurvey}. What is worse, it has been shown that federated models are more unreliable on Non-IID data than centralized training models, demonstrating severe miscalibrations on in-distribution data and low uncertainty on OOD data \citep{APH}. Such a nature poses significant challenges to the reliability and explainability of federated models, making the construction of trustworthy federated learning frameworks harder. To the best of our knowledge, the trustworthy federated learning framework, which simultaneously possesses robustness, security, reliability, and high performance, remains to be established.

To bridge this gap, we propose Trustworthy Personalized Federated Learning (TPFL), a secure, robust, efficient, and reliable federated framework with high performance on Non-IID data. Specifically, TPFL integrates subjective logic (SL) into the training schema, enabling single-forward uncertainty estimation within inference. In the local training stage, TPFL formulates both global and local models in a subjective manner, forcing them to give subjective opinions with uncertainty on instances instead of direct first-order probability. To further combat data heterogeneity, a trainable heterogeneity prior to opinion formulation and re-balance fine-tuning personalized strategy has been introduced to counter the impact of class imbalance and distribution skew. Global and local opinions are then fused under the guidance of uncertainty to give more comprehensive and accurate decisions. In aggregation, TPFL further estimates the model uncertainty of each uploaded model in the server. Model uncertainty is further utilized to judge and filter suspected malicious updates. Empirically, we conduct extensive evaluations on various popular federated benchmarks, including Cifar10, Cifar100, and Tiny-ImageNet. Experimental results demonstrate that TPFL not only surpasses most state-of-the-art (SOTA) methods on model performance but also achieves robustness against domain shifts, reliability on high-stakes scenarios, and safety against most advanced malicious attacks.

\underline{\textbf{Technical Contributions:}} In this paper, we take a significant step towards establishing a trustworthy FL framework. We summarize our contributions as below:
\begin{itemize}
    \item We first integrate subjective logic into the federated learning framework and establish Trustworthy Personalized Federated Learning (TPFL), which can provide predictive uncertainty to assist manual decisions and refuse to predict when meeting OOD data.
    \item We propose the first effective but efficient uncertainty-based server-side anomaly detection strategy, effectively encountering malicious attacks with a small computational cost.
    \item We conduct extensive experiments on various popular federated benchmarks, validating the safety, robustness, reliability, and superior performance of TPFL.  
\end{itemize}
\section{Related Work}
\textbf{Federated Learning on Non-IID Data}: Federated learning (FL) has become a promising decentralized training method since FedAvg's introduction \cite{FedAvg}, but struggles with performance and convergence on heterogeneous data \cite{Exp_NonIID}. Generalized FL (G-FL) methods address this by mitigating client drift \cite{FedProx, MOON, FedDC}, applying gradient corrections \cite{SCAFFOLD, FedAvgM, FedDyn, GradMA}, or using knowledge distillation \cite{FedDF, FedNTD, FedKA}. Personalized FL (P-FL), first introduced by \cite{FedMTL}, aims to customize models for each client and includes strategies such as model splitting \cite{FedRep, FedRoD}, model mixture \cite{pFedMe, ditto, GPFL}, personalized aggregation \cite{FedFOMO, FedAMP, FedALA}, and clustering \cite{ClusterFL, HRCFL}. Though achieving superior performance, most current P-FL methods lack the comprehensive fusion of judgments from both generic and personalized models, significantly affecting the precision and reliability of predictive decisions.

\textbf{Safety in FL}: Safety in FL aims to defend against adversarial attacks, both targeted (backdoors) \cite{BackdoorFL, TargetedBackdoor} and untargeted (model degradation) \cite{PoisonEvaluation}. Untargeted attacks can involve data poisoning \cite{FlipAttack} or model poisoning \cite{ModelReplacement, LocalModelPoisoning, LIE, PoisonEvaluation, MPAF}. Defense methods include robust aggregation, using median or trimmed means \cite{MedianMeanTrMedian, RobustAggregation, Krum}, and anomaly detection \cite{RobustCFL, Biscotti, BlockChainSecurity}. Although great success has been achieved, balancing safety and performance remains challenging.

\textbf{Reliability in FL}: Reliability in FL refers to well-calibrated models with uncertainty estimates \cite{EDL, APH, ECE}. Recent studies reveal that models trained on non-IID data are unreliable \cite{APH}.  Motivated by this, numerous works have been endeavored to transfer traditional uncertainty estimation methods \cite{Dropout, DeepEnsemble} to federated scenarios \citep{PioneerUncertaintyFL, FedEnsemble}, or try to propose novel methods \citep{FCP, APH}. However, these methods often incur high computation costs.

\textbf{Subjective Logic (SL)}: SL, a generalized form of Dempster-Shafer Theory (DST) \cite{DST, DSTBook}, introduces epistemic uncertainty through belief masses and allows evidence fusion via combination rules \cite{SLrule0, SLrule1}. SL has been applied to deep learning for uncertainty quantification \cite{EDL}, and recently to FL via DST in RIPFL \cite{RIPFL}. However, the disregard for prior distributions of DST not only leads to the incompleteness of the applied theory \cite{SubjectiveLogic} but is also unacceptable in practical federated scenarios, particularly in Non-IID settings where data distributions are heterogeneous \cite{Exp_NonIID} among all clients.

A detailed description of related work is in the Appendix.

\section{Trustworthy Personalized Federated Learning via Subjective Logic}
In this section, we introduce our proposed trustworthy personalized federated learning (TPFL), which integrates subjective logic into the decentralized training and utilizes instance and model uncertainty to combat heterogeneity with adversarial clients. To introduce the framework, we decouple the whole algorithm into four parts, in which we will describe the background of subjective models, the local training, the inference procedure, and the aggregation stage.

\begin{figure*}
    \centering
    \includegraphics[width=0.98\linewidth]{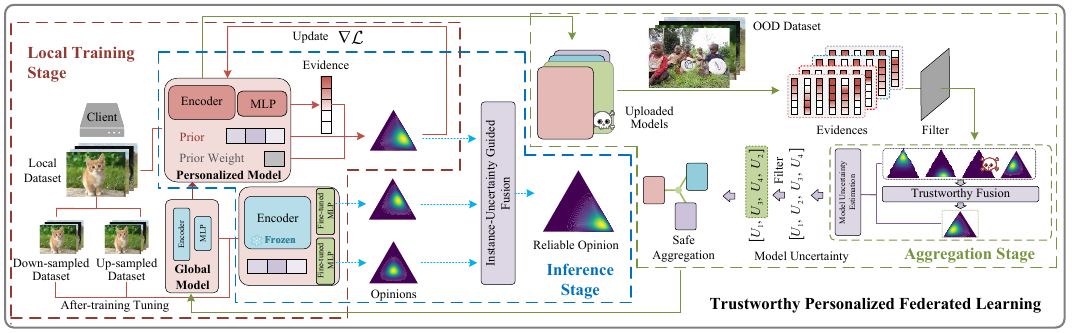}
    \caption{Overall framework of Trustworthy Personalized Federated Learning (TPFL). In the local training stage (red part), data is directly utilized to train the subjective model and adjust its local prior distribution (See section \ref{sec3.2} for detailed loss designs). The local model is further uploaded to the central server for aggregation (green part). To ensure training safety, TPFL utilizes an out-of-distribution (OOD) dataset with a small size to evaluate each uploaded model. Leveraging the sensitive property of subjective models (See section \ref{sec3.2}), evidences are first filtered to exclude updates with extremely large or overflow values, protecting the aggregated model from attacks utilizing significant tempering \cite{ModelReplacement}. The remaining opinions are further integrated via instance uncertainty-guided trustworthy fusion. Discrepancies between original and fused opinions are treated as model uncertainty to measure the similarity of updates. Too similar updates with small model uncertainty deviation will be filtered to counter accumulation attacks \cite{LIE, LocalModelPoisoning, MPAF}. Finishing all training, the local models will be required to fine-tune on re-balanced datasets, and the inference (blue part) will integrate three sources to produce reliable predictions.}
    \label{fig_sm}
\end{figure*}

\subsection{Forming Opinions via Subjective Logic}
\label{sec3.1}


Subjective logic is an uncertain probabilistic logic introduced by \citet{SubjectiveLogic}, relating prior, belief and uncertainty with probabilities to form opinions for decision-making. By establishing the mapping between multinomial opinion and Dirichlet PDF, the subjective opinion, which is determined by belief masses, prior, and uncertainty, can be easily transformed into a probability with uncertainty, providing a trustworthy manner for prediction \cite{SLReasoning}. Motivated by the untrustworthy nature of current federated frameworks, we integrate subjective logic into the decentralized training schema of personalized FL, which aims to utilize inherent instance uncertainty to fuse valuable information between personalized and generic opinions of each sample to improve model performance and exploit estimated model uncertainty to scrutinize malicious updates to ensure training safety.

Considering a horizontal federated scenario with Non-IID data, PFL aims to learn a personalized model for each client, mitigating the inconsistency of optimization objectives caused by label distribution skew. Unlike deterministic models, we refactor deployed models subjectively, forcing them to observe evidence for each possible state to provide subjective opinions with uncertainty rather than directly estimating probabilities.

\begin{definition}[Multinomial Opinion of Subjective Federated Model]
For each sample ${X} \in \mathbb{X}$, let ${Y}$ be the random variable of its label over the domain $\mathbb{Y}$, where $|\mathbb{Y}| = k $  for a $k$-classification problem. Given the prior distribution $\boldsymbol{a}_{Y}^{i}\text{ s.t. } \sum_{y \in \mathbb{Y}}{\boldsymbol{a}_{Y}^{i}(y)} = 1$ of class assignments, the personalized multinomial opinion $\omega_{Y}^i$ of client $c_i$ is defined as:
\begin{equation}
    {\omega}_{Y}^i \coloneqq (\boldsymbol{b}_Y^i, u_Y^i, \boldsymbol{a}_Y^i) 
\end{equation}
, where $\boldsymbol{b}_{Y}^i : \mathbb{Y} \rightarrow [0, 1]$ is the belief mass distribution, $u_Y$ is the uncertainty mass with requirement:
\begin{equation}
    u_Y^i + \sum_{y \in \mathbb{Y}}{\boldsymbol{b}_{Y}(y)} = 1
\label{eqcalu}
\end{equation}

Similarly, the global opinion $\omega_{Y}^i$ is defined with similar requirements as:
$
    {\omega}_{Y}^g \coloneqq (\boldsymbol{b}_Y^g, u_Y^g, \boldsymbol{a}_Y^g)  
$
\end{definition}

Instead of directly providing the probability distribution of label variable $Y$, the subjective federated model gives a more precise opinion via an ordered triple, assigning belief mass to each singleton value $y \in \mathbb{Y}$ and forming the uncertainty on this opinion. Following the guidance of subjective logic, we further relate subjective opinions to Dirichlet distributions through a bijective mapping \cite{SubjectiveLogic}.

\begin{definition}[Bijective Mapping Between Multinomial Opinion and Dirichlet PDF] For each multinomial opinion ${\omega}_{Y} = (\boldsymbol{b}_Y, u_Y, \boldsymbol{a}_Y) $, there exists a unique corresponding Dirichlet distribution $\operatorname{Dir}(\boldsymbol{\alpha}_Y)$, in which:
\begin{equation}
    \boldsymbol{\alpha}_Y = \boldsymbol{e}_Y + W\boldsymbol{a}_{Y}
\end{equation}. $W$ is the non-informative prior weight and $\boldsymbol{e}_Y$ is the evidences related to the belief masses $\boldsymbol{b}_Y$:
\begin{equation}
    \boldsymbol{b}_Y = \frac{\boldsymbol{e}_Y}{W + \sum\limits_{y \in \mathbb{Y}}{\boldsymbol{e}_Y} (y)} 
\label{eqcalb}
\end{equation}
\end{definition}

Intuitively, $\boldsymbol{\alpha}$ is decided by the observed evidence $\boldsymbol{e}$ and the attached term $W\boldsymbol{a}$. $W$ is a preset value that determines the weight of prior knowledge, always setting to the class number $k$ in practice. The term $W \boldsymbol{a}$ is called basic evidence,  which provides the basic value of $\boldsymbol{\alpha}_Y$ even if no corresponding evidences are observed. Such a mapping establishes a direct mathematical interpretation on the subjective opinion, allowing us to build a probability density distribution of label random variable $Y$ over the domain $\mathbb{Y}$. Thus, the expectation probability $\mathbb{P}_{\hat{y}}$ on the singleton value $\hat y \in \mathbb{Y}$ can be easily obtained through expectation: 
\begin{equation}\label{eqmap}
    \mathbb{P}_{\hat{y}} = \mathbb{E}\left[\hat{y}\right]=\frac{\boldsymbol{\alpha}_{Y}({\hat{y}})}{\sum_{y \in \mathbb{Y}}{\boldsymbol{\alpha}_Y(y)}} = \boldsymbol{b}_Y(\hat{y})+\boldsymbol{a}_Y(\hat{y}) u_Y
\end{equation}

We visualize the overall inference process of the subjective model in Fig.\ref{fig1}. In brief, the subjective model preserves instance-irrelevant prior distribution $\boldsymbol{a}$ and prior weight $W$. Given a sample $X$, it outputs observed evidence $\boldsymbol{e}$, which are further transformed into belief masses $\boldsymbol{b}$ on singletons with uncertainty $u$ via Eq.[\ref{eqcalb}] and Eq.[\ref{eqcalu}]. Subsequently, the triplet $(\boldsymbol{b}, u, \boldsymbol{a})$ forms reliable subjective opinions, which can be further calculated into probability assignments through bijective mapping and expectation.

\subsection{Robust Local Training with Trainable Heterogeneity Prior}
\label{sec3.2}

Inspired by the previous work \cite{EDL, Red} based on the DS evidence theory \citep{DST, DSTBook}, we further extend the training framework to subjective models, allowing informative prior integration to guide the model judgments. To the best of our knowledge, we are the first to establish the framework of prior-involved subjective model training.

 We reclaim that the purpose of subjective models is to collect observed evidence for each sample rather than directly give out their predictions. So, it is intuitive to require the evidence $\boldsymbol{e}$ to be non-negative. Technically, we choose a proper activation function (e.g., ReLU \cite{ReLU}, Exponential) to ensure the validity of observed evidence and further remove the conventional Softmax layer to avoid normalization as done in \citep{EDL}. Considering potential malicious attacks involved in collaborative training, we choose the Exponential function to effectively counter the scaling of model parameters launched by model replacement attack \citep{ModelReplacement} (See more discussions in Experiments section). 

Since the model's output is observed evidence, commonly used loss functions (e.g., cross-entropy loss) are unsuitable for subjective models. Thus, we adopt a variant of cross-entropy loss proposed in \citep{EDL} to train the subjective models:
\begin{equation}
    \mathcal{L}_{\text{CE}}(\mathbf{x}, \mathbf{y})=\sum_{j=1}^K y_k\left(\Psi(S)-\Psi\left(\alpha_k\right)\right)
\end{equation}

Solely requiring the subjective model to make correct answers is insufficient. To ensure its trustworthy property, it is also important to regularize subjective models to avoid observing incorrect evidence. To this end, we design a regularization loss for subjective models based on Kullback-Leibler divergence following \citep{EDL, Red}:
\begin{equation}
\label{KL}
\begin{split}
    \mathcal{L}_{\text{inc}} &= \operatorname{KL}\left(\operatorname{Dir}\left(\boldsymbol{p} \mid \boldsymbol{\hat{\alpha}}\right) \| \operatorname{Dir}\left(\boldsymbol{p} \mid \boldsymbol{a}\right)\right) \\
    &= \ln \frac{\Gamma\left(\sum_{i=1}^k \hat\alpha_{ i}\right)}{\Gamma\left(\sum_{i=1}^k a_{ i}\right)}+ \sum_{i=1}^k \ln \frac{\Gamma\left(a_{ i}\right)}{\Gamma\left(\hat\alpha_{ i}\right)} \\
    & +\sum_{i=1}^k\left(\hat\alpha_{i}-a_{i}\right) \left[\psi\left(\hat\alpha_{ i}\right)-\psi\left(\sum_{i=1}^k \hat\alpha_{ i}\right)\right]
\end{split}
\end{equation}
, where $\boldsymbol{\hat\alpha} = \left(1-\boldsymbol{y}\right) \odot \boldsymbol{\alpha} +  W\boldsymbol{y} \odot\boldsymbol{a}$ is the trimmed concentration parameter of Dirichlet distribution, $\Gamma(\cdot)$ is the gamma function and $\psi(\cdot)$is the digamma function. The $\hat\alpha_{i=gt}$, which corresponds to the true label, is trimmed to its basic evidence while the others are retained. Thus, minimizing the KL divergence between two Dirichlet distributions $\operatorname{Dir}\left(\boldsymbol{p} \mid \boldsymbol{\hat{\alpha}}\right) $ and $\operatorname{Dir}\left(\boldsymbol{p} \mid \boldsymbol{a}\right)$ is to suppress erroneous evidence observations and wish observed evidence on the non-ground-truth label to be zero. Though intuitive and effective, such a regularization may cause the zero-evidence regions \citep{Red} on hard samples. To encourage the observation of evidence on zero-evidence regions, we utilize a correct evidence encouragement term based on \citep{Red}:
\begin{equation}
    \mathcal{L}_{\text {cor }}(\boldsymbol{x}, \boldsymbol{y})=-u\ln \left(\alpha_{g t}-a_{gt}\right)
\end{equation}

Besides, we observe that skewed local datasets in practical Non-IID federated data scenarios can easily lead to gradient explosion with infinite evidence in majority classes. To suppress that, we further design a regularization term to restrict extremely large or even overflow evidence values:
\begin{equation}
\mathcal{L}_{\text{evi}} = \| \operatorname{max}\left(\boldsymbol{0}, \boldsymbol{e} - \epsilon \right) \|^2_2
\label{evi_reg}
\end{equation}, where $\epsilon$ is the evidence threshold set to 10000 in our experiments.

To further leverage previous training knowledge and handle the skewed data distribution, we replace the static prior with a trainable one, enabling it to self-adapt according to the past training. Such a design allows the same model to perform differently and adaptively according to its deployed data distribution, significantly loosening the constraints between skewed clients. Note that Eq.\ref{KL} will not be used to update local prior. Normalization will be conducted after updating. We utilize the following term to ensure the non-negative property of prior distributions:
\begin{equation}
    \mathcal{L}_{\text{neg}} = \| \boldsymbol{a} - \operatorname{max}(\boldsymbol{0}, \boldsymbol{a}) \|_1
\end{equation}

The total loss of local training is calculated as follows:
\begin{equation}
    \mathcal{L} = \mathcal{L}_{\text{CE}} + 
 \mathcal{L}_{\text{cor}}
    +
    \lambda_1 \mathcal{L}_{\text{inc}}  +
    \lambda_2
    \mathcal{L}_{\text{evi}} +
    \lambda_3
    \mathcal{L}_{\text{neg}}
\end{equation}
where $\lambda_1$, $\lambda_2$, and $\lambda_3$ are coefficients to control the balance of loss functions. Note that the latter two terms are utilized to ensure the numerical stability of the framework. Thus, our proposed TPFL are insensitive to their values. More discussions about the hyperparameters can be found in experiments. TPFL also applies a model-split personalization strategy in clients, which only replaces the feature encoder with the generic one and keeps the classifier unchanged at the beginning of each communication round.



\subsection{Debiased Trustworthy Inference via Instance Uncertainty-Informed Opinion Fusion}

With subjective models introduced, we carefully describe the debiased trustworthy inference, with the instance uncertainty-informed opinion fusion. 

The design of debiased trustworthy inference is motivated by critical observations when we compare the judgments between generic and personalized models: There exists a large number of 'p-hard' instances that can be easily classified correctly by the generic model but misclassified by personalized models (See more observations and exploration experiments details in Appendix). Such an observation indicates that the personalized and generic information is complementary rather than replaceable, calling for the effective integration of personalized and generic information in inference to improve performance.

Motivated by this, we propose utilizing instance uncertainty with knowledgeable prior to effectively integrating generic and personalized opinions in a trustworthy manner. As discovered in \citep{FedETF, APH}, classifiers in generic models are always biased due to imbalanced data, posing obstacles to their trustworthy cooperation with personalized models. So, during inferences in TPFL, instead of directly utilizing biased generic opinions, we design a lightweight post-hoc strategy to integrate generic and personalized information effectively. Concretely, after all the training, we adopt a re-balance strategy to fine-tune the classifiers of subjective generic models on manually balanced local data, eliminating the biases introduced by skewed aggregation. For local dataset $\mathcal{D}_i$, we utilize both up-sample and down-sample strategies to get the balanced local datasets $\mathcal{D}_i^u$ and $\mathcal{D}_i^d$ respectively. A few-round fine-tuning on projection layers is further conducted to obtain the debiased generic model $\theta_{g,u}$ and $\theta_{g,d}$ with frozen encoders. Note that balance-tuning only happens once for each client after all the training. 

During inference, each test sample will receive three opinions from different sources. We measure the reliability of these opinions by instance uncertainty and further leverage instance uncertainty-informed opinion fusion to get reliable aggregated opinions for judgment.

\begin{definition}(Instance Uncertainty-Informed Opinion Fusion) Given two opinions $\omega^{A}_{Y} = (\boldsymbol{b}^{A}_{Y}, u^{A}_{Y}, \boldsymbol{a}^{A}_{Y})$ and $\omega^{B}_{Y} = (\boldsymbol{b}^{B}_{Y}, u^{B}_{Y}, \boldsymbol{a}^{B}_{Y})$ from reliable sources, the instance uncertainty-informed aggregated opinion $\omega^{(A \widehat{\diamond} B)}_{Y} = 
\omega_Y^A  \oplus  \omega_Y^B
= (\boldsymbol{b}_{Y}^{A \widehat{\diamond} B}, u_{Y}^{A \widehat{\diamond} B}, \boldsymbol{a}_{Y}^{A \widehat{\diamond} B})$ is calculated as:

\begin{equation}
\left\{\begin{aligned}
\boldsymbol{b}_{Y}^{A \widehat{\diamond} B} & =\frac{\boldsymbol{b}_{Y}^A\left(1-u_{Y}^A\right) u_{Y}^B+\boldsymbol{b}_{Y}^B\left(1-u_{Y}^B\right) u_{Y}^A}{u_{Y}^A+u_{Y}^B-2 u_{Y}^A u_{Y}^B} \\
u_{Y}^{A \widehat{\diamond} B} & =\frac{\left(2-u_{Y}^A-u_{Y}^B\right) u_{Y}^A u_{Y}^B}{u_{Y}^A+u_{Y}^B-2 u_{Y}^A u_{Y}^B}, \\
\boldsymbol{a}_{Y}^{A \widehat{\diamond} B} & =\frac{\boldsymbol{a}_{Y}^A\left(1-u_{Y}^A\right)+\boldsymbol{a}_{Y}^B\left(1-u_{Y}^B\right)}{2-u_{Y}^A-u_{Y}^B},
\end{aligned}\right.
\end{equation}

\end{definition}

Intuitively, instance uncertainty-informed opinion fusion measures the confidence of provided opinions through uncertainty and further utilizes them to guide the fusion of opinions from different sources. We further demonstrate that the aggregation through instance uncertainty-informed fusion exactly means the weighted average of opinion evidence based on instance uncertainty and prior (See Appendix). Let $\omega_Y^{g,u}$ and $\omega_Y^{g,d}$ denote the opinions from the up-debiased model and down-debiased model, respectively, the final aggregated opinion $
    \omega_Y = \omega_Y^{g,u} \oplus \omega_Y^{g,d} \oplus \omega_Y^{p}$, 
can be easily mapped to a unique Dirichlet distribution and get probability distribution through expectation in Eq. \ref{eqmap}.



\subsection{Secure Aggregation with Model Uncertainty-based Anomaly Detection}
As indicated before, malicious attacks greatly affected the training and safety of federated frameworks. To ensure the trustworthiness of TPFL, we design a secure aggregation module with model uncertainty-based anomaly detection, effectively filtering the abnormal updates. Malicious attacks, whether targeted or untargeted, all aim to replace the good global model with a corrupted one. Such a replacement can be conducted through two manners: (1) dramatic scaling on corrupted model parameters to increase their aggregation weight \cite{ModelReplacement} implicitly; (2) fault accumulation via multiple updates with similar but trivial tampering \cite{LIE, LocalModelPoisoning}. 
To protect federated frameworks from attacks, think of sending and recollecting boxes. One approach is to make the boxes extremely sturdy, preventing any tampering. Another is to make them so fragile that even the slightest interference causes them to break, exposing any manipulation. Similarly, the safety of TPFL comes from its two properties: robustness and sensitivity. 

On the one hand, TPFL employs separate shared encoders and personalized classifiers, ensuring that minor but unorganized tampering does not affect the benign models. On the other hand, the sensitivity of TPFL makes it easy to detect significant scaling or similar updates with trivial but organized tampering. Specifically, as the activation function of subjective models chosen as exponential, scaling or significantly tampering parameters of a benign model can easily lead to extremely large evidence output or overflow, resulting in easy filtering on these malicious updates. Besides, for similar updates with trivial but organized tampering, we extend the instance uncertainty estimation to the model level (defined below), enabling the numerical evaluation of uploaded model parameters.

\begin{definition}(Model Uncertainty) Let $\theta_i$ denote the personalized model of client $c_i$, $\mathcal{D}^{\text{h}}$ denote the trusted OOD holdout validation dataset possessed in the central server, the model uncertainty of client $c_i$ is defined as:
\begin{equation}
\footnotesize
U(\theta_i ; \mathcal{D}^{\text{h}}) = \frac{1}{\|\mathcal{D}^{\text{h}}\|}\sum_{X_j \in \mathcal{D}^{\text{h}}}\operatorname{KL}\left(\operatorname{Dir}\left(\boldsymbol{p}\mid\alpha_{Y_j} \right)\|\operatorname{Dir}\left(\boldsymbol{p}\mid \bar\alpha_{Y_j}\right)\right) 
\end{equation}
, where $\operatorname{Dir}\left(\boldsymbol{p}\mid \bar\alpha_{Y_j}\right)$ is the mapped Dirichlet distribution of aggregated opinions among filtered clients.

\end{definition}

Model uncertainty $U$ measures the discrepancy severity of model opinions on the public trusted validation set between the aggregated opinions, enabling the mapping from opinions on the whole dataset to a precise numerical value. Such a transformation is sensitive, explicitly reducing the dimension of model parameters to one. So when values of model uncertainty are highly similar, it indicates nearly complete overlap between their opinions on the whole holdout datasets, which conforms to a critical feature of organized attacks with trivial tampering aiming to accumulate faults. Thus, such updates with similar model uncertainty will also be filtered from model aggregation. The overall pseudo-code of TPFL can be found in the Appendix.

\begin{table*}[t!]
  \centering
  \caption{Predictive accuracy comparison of TPFL with other SOTA methods in various federated settings. }
  \resizebox{0.75\textwidth}{!}{
    \begin{tabular}{cccccccc}
    \toprule
    \multirow{3}[6]{*}{Method} & \multicolumn{5}{c}{Cifar10}           & Cifar100 & Tiny-ImageNet \\
\cmidrule{2-8}          & Client 10 & Client10 & Client 20 & Client 100 & Client 10 & Client 10 & Client 10 \\
\cmidrule{2-8}          & $\beta$=0.05 & $\beta$=0.1 & $\beta$=0.1 & $\beta$=0.1 & $\beta$=0.5 & $\beta$=0.1 & $\beta$=0.1 \\
    \midrule
    FedAvg \cite{FedAvg} & 0.584±0.003 & 0.609±0.004 & 0.585±0.002 & 0.557±0.009 & 0.674±0.006 & 0.291±0.004 & 0.110±0.002 \\
    FedProx \cite{FedProx} & 0.535±0.006 & 0.563±0.005 & 0.516±0.001 & 0.427±0.001 & 0.650±0.005 & 0.191±0.002 & 0.071±0.003 \\
    MOON \cite{MOON} & 0.327±0.007 & 0.497±0.001 & 0.318±0.013 & 0.352±0.003 & 0.556±0.006 & 0.284±0.002 & 0.095±0.004 \\
    \midrule
    Per-FedAvg \cite{Per-FedAvg} & 0.916±0.001 & 0.826±0.008 & 0.856±0.004 & 0.826±0.009 & 0.682±0.009 & 0.298±0.002 & 0.127±0.001 \\
    FedRep \cite{FedRep} & 0.943±0.001 & 0.895±0.000 & 0.897±0.001 & 0.873±0.001 & 0.757±0.003 & 0.430±0.004 & 0.264±0.009 \\
    pFedMe \cite{pFedMe} & 0.941±0.001 & 0.892±0.002 & 0.896±0.002 & 0.813±0.002 & 0.749±0.003 & 0.412±0.001 & 0.279±0.002 \\
    Ditto \cite{ditto} & 0.943±0.001 & 0.893±0.002 & 0.898±0.001 & 0.868±0.001 & 0.772±0.001 & 0.439±0.004 & 0.329±0.003 \\
    FedRoD \cite{FedRoD} & 0.935±0.002 & 0.878±0.002 & 0.898±0.002 & 0.882±0.001 & 0.757±0.007 & 0.391±0.008 & 0.233±0.006 \\
    FedFomo \cite{FedFOMO} & 0.933±0.001 & 0.882±0.001 & 0.883±0.001 & 0.863±0.002 & 0.692±0.001 & 0.382±0.001 & 0.254±0.003 \\
    APPLE \cite{APPLE}& 0.938±0.002 & 0.891±0.002 & 0.894±0.003 & -     & 0.767±0.003 & 0.423±0.001 & 0.327±0.002 \\
    FedALA \cite{FedALA} & 0.938±0.001 & 0.886±0.001 & 0.900±0.004 & -     & 0.749±0.001 & 0.313±0.002 & 0.239±0.002 \\
    TPFL  & 0.944±0.002 & 0.901±0.003 & 0.907±0.002 & {0.883±0.001} & 0.751±0.003 & 0.442±0.004 & 0.293±0.004 \\
    TPFL-0.5 & 0.946±0.001 & 0.902±0.001 & 0.907±0.003 & {0.884±0.001} & 0.767±0.002 & 0.507±0.005 & 0.449±0.002 \\
    TPFL-0.2 & \textbf{0.951±0.001} & \textbf{0.920±0.002} & \textbf{0.920±0.002} & \textbf{0.898±0.001} & \textbf{0.822±0.003} & \textbf{0.874±0.003} & \textbf{0.588±0.003} \\
    \bottomrule
    \end{tabular}%
    }
  \label{predictive_acc}%
\end{table*}%

\section{Experiments}
We conducted extensive experiments to validate the effectiveness of TPFL from various perspectives. The experimental setup and more results are in the Appendix.



\subsection{Predictive Performance}
We compare the performance of TPFL in Non-IID data generated by Dirichlet distribution $\operatorname{Dir}(\beta)$ with different SOTA federated methods. For a fair comparison, we train all the methods three times and report the mean with variance. Results are displayed in Table.\ref{predictive_acc}. As indicated in the table, even without the assistance of instance uncertainty, TPFL consistently achieves high performance on popular federated benchmarks with different sizes and severity levels of data heterogeneity, surpassing most existing personalized methods. When treating instance uncertainty as the decision threshold, TPFL can perform better. For example, TPFL-0.5, which means the model rejects making a decision when uncertainty reaches 0.5, can achieve up to 20\% improvement. We also report the accuracy curve to indicate the fast convergence of TPFL in the Appendix.

\begin{figure}[t!]
\captionsetup[subfigure]{justification=centering}
    \centering
      \begin{subfigure}{0.18\textwidth}
        \includegraphics[width=\textwidth]{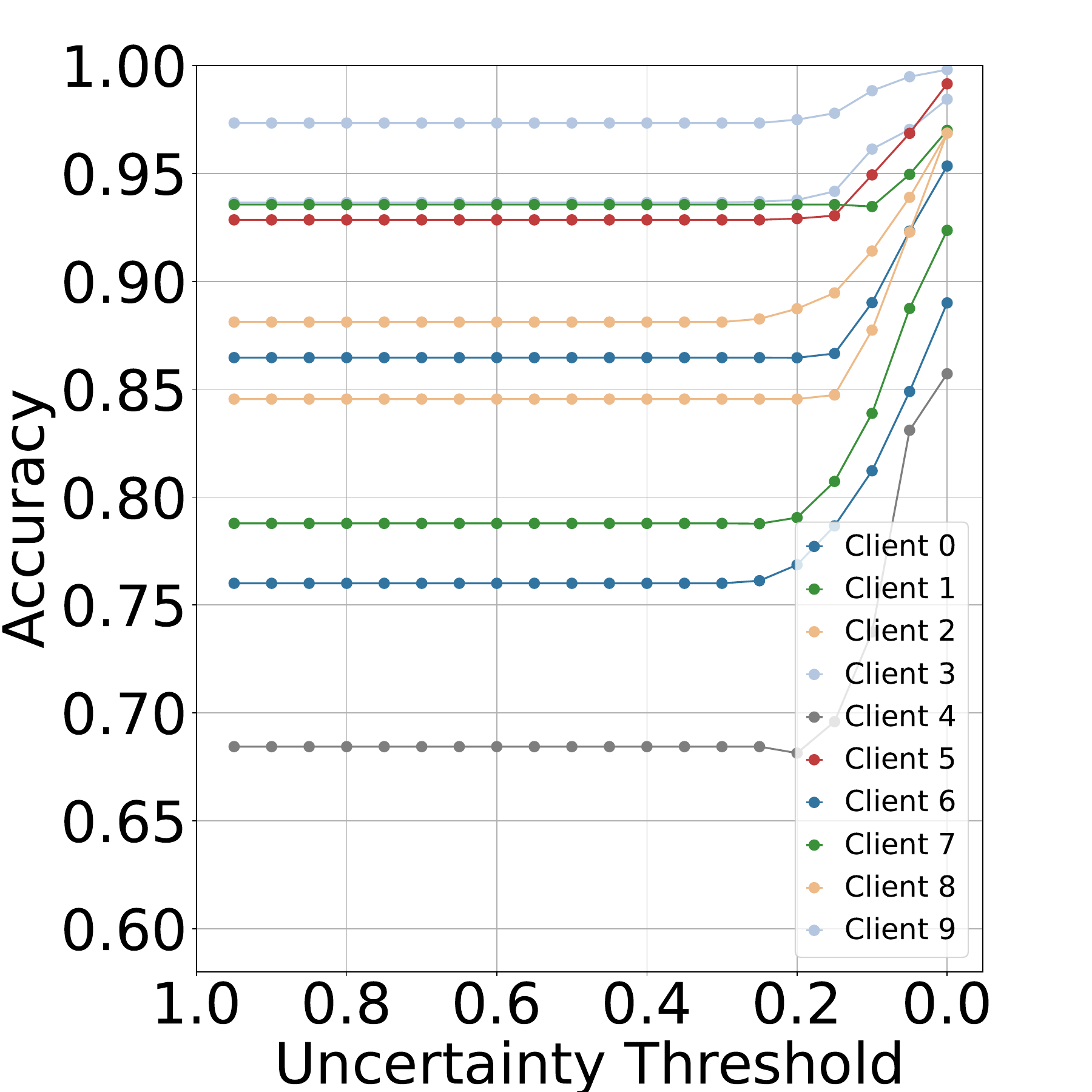}
          \caption{In-distribution Cifar10}
          \label{ra}
      \end{subfigure}
            \begin{subfigure}{0.18\textwidth}
        \includegraphics[width=\textwidth]{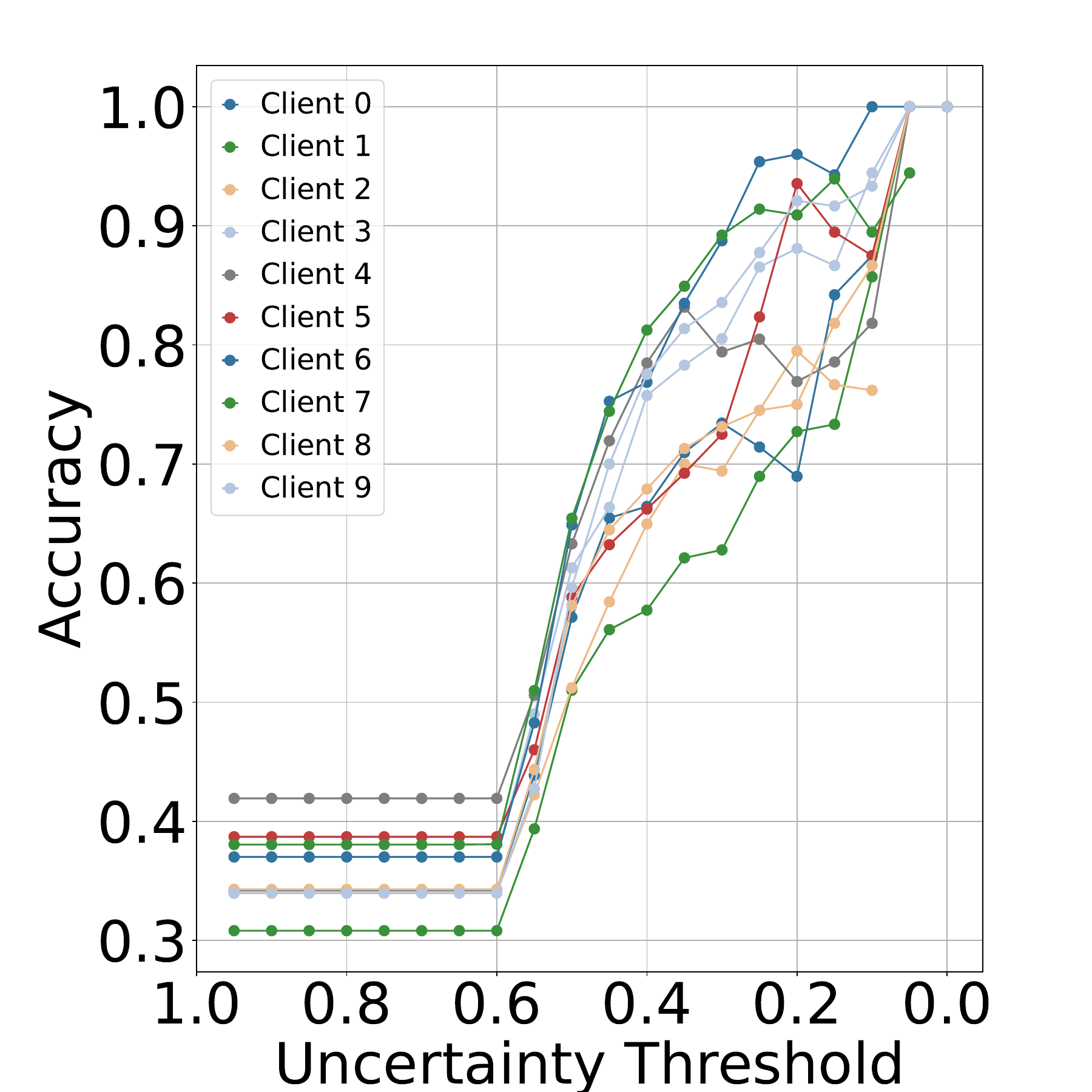}
          \caption{In-distribution Cifar100}
          \label{rb}
      \end{subfigure}
          \begin{subfigure}{0.18\textwidth}
        \includegraphics[width=\textwidth]{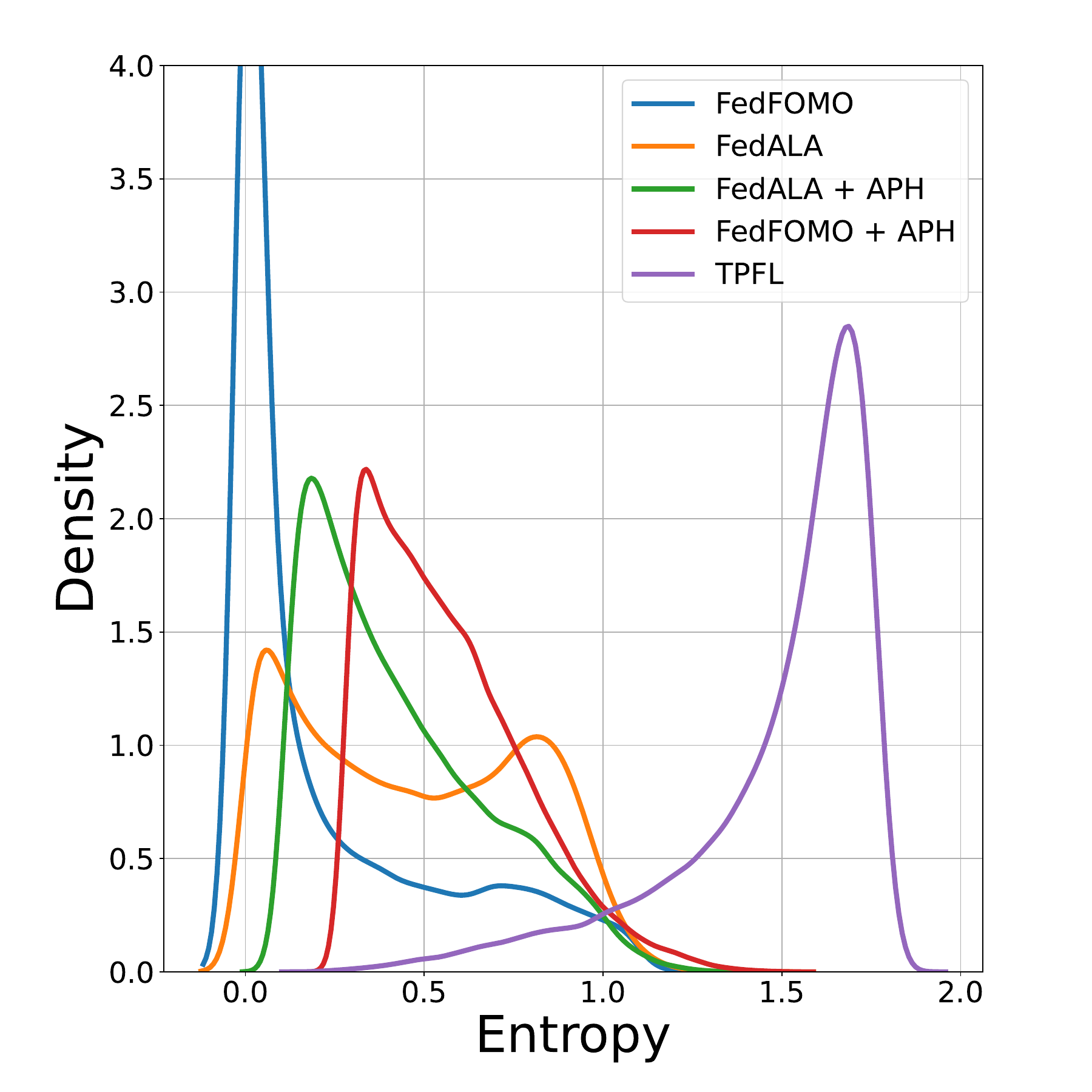}
          \caption{SVHN Vs. Cifar10}
          \label{rc}
      \end{subfigure}
        \begin{subfigure}{0.18\textwidth}
    \includegraphics[width=\textwidth]{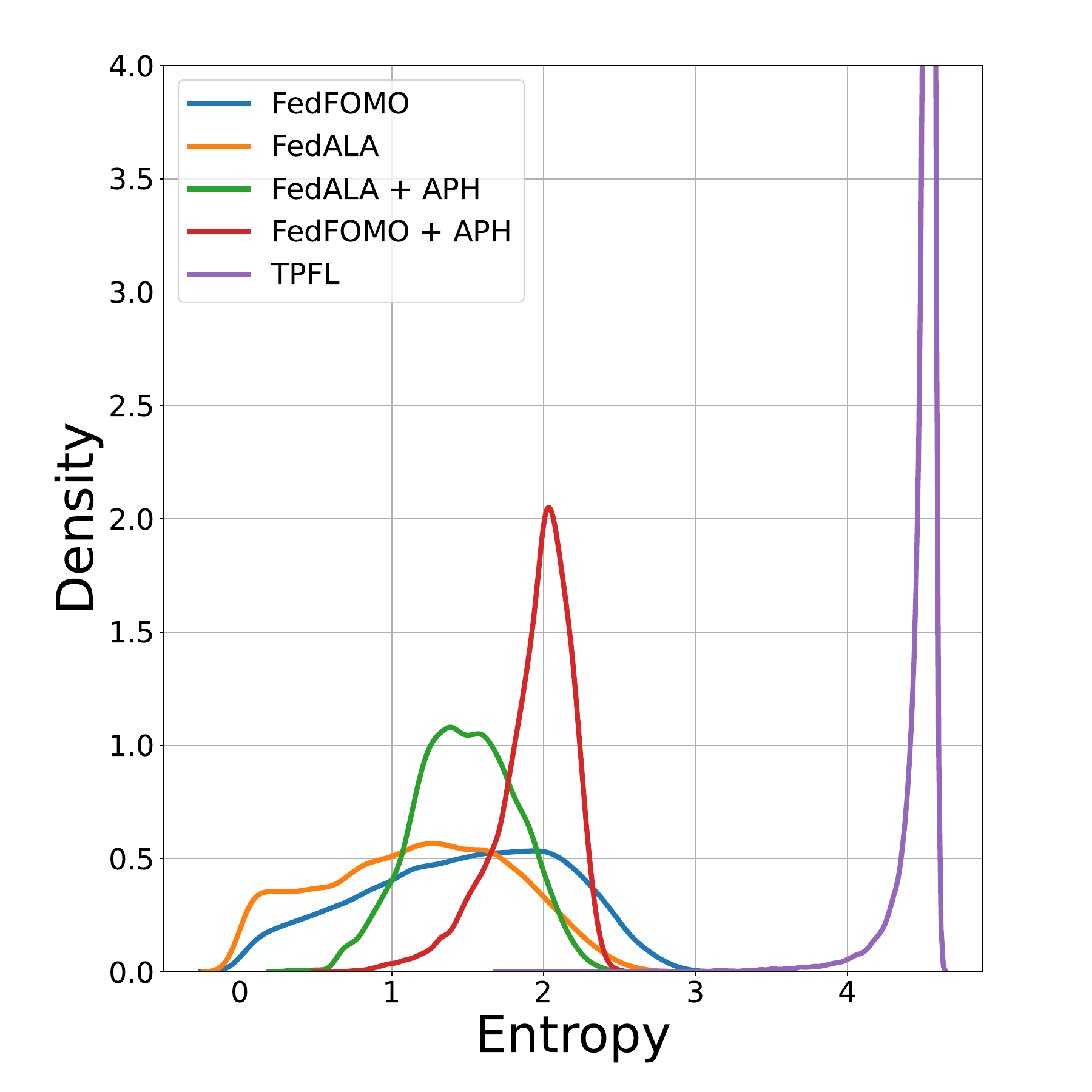}
          \caption{SVHN Vs. Cifar100}
          \label{rd}
      \end{subfigure}
    \caption{Reliability experimental results. Fig.\ref{ra} and \ref{rb} demonstrate the effectiveness of estimated uncertainty. With the decrease of the threshold, the predictive accuracy significantly increases. Fig.\ref{rc} and \ref{rd} validate the reliability of TPFL on OOD datasets, showcasing higher uncertainty compared with other methods. }
    \label{Reliability}
\end{figure}

\begin{figure*}[t!]
    \captionsetup[subfigure]{justification=centering}
    \centering
    \begin{subfigure}{0.15\textwidth}
\includegraphics[width=\textwidth]{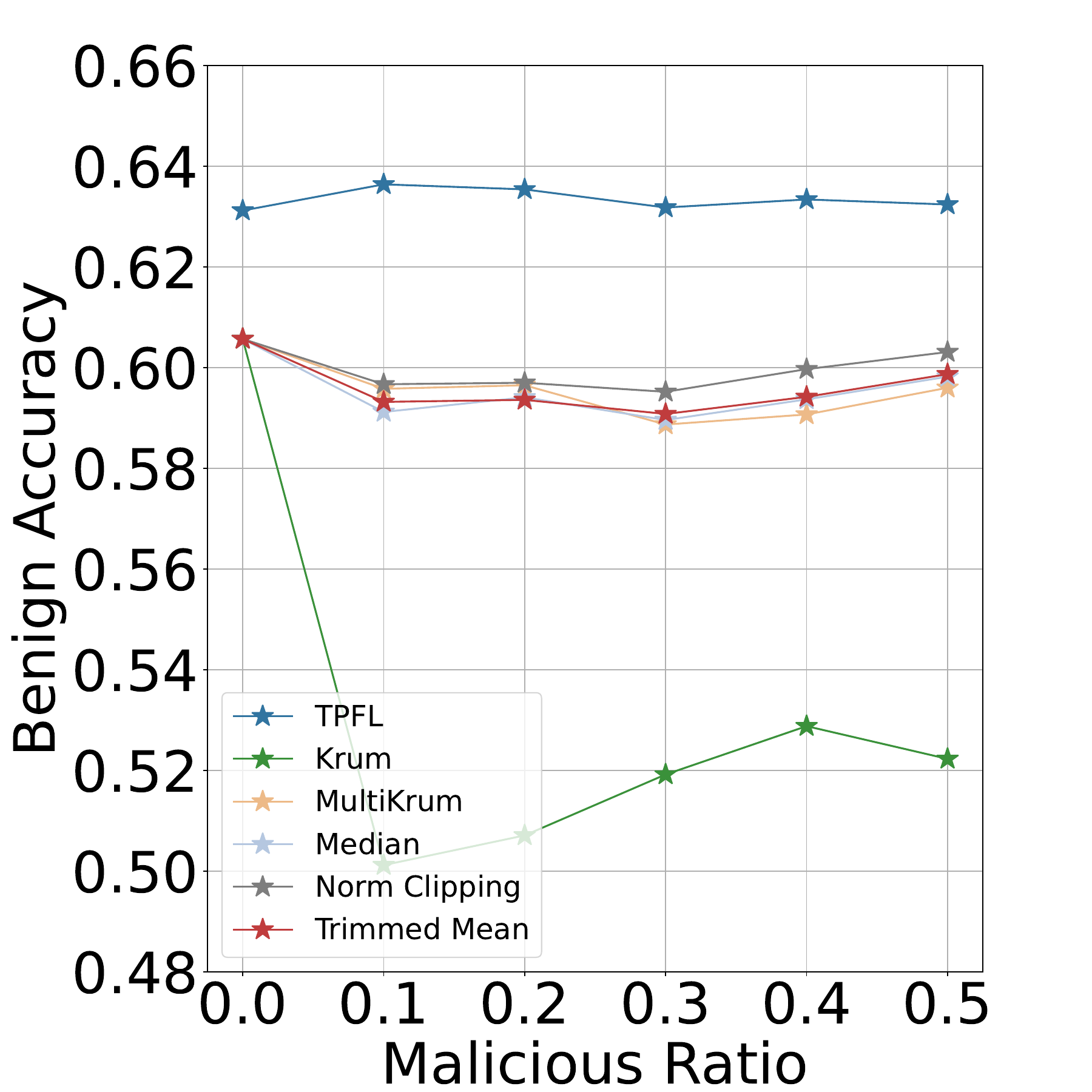}
          \caption{Label Flip \cite{FlipAttack}}
          \label{sa}
    \end{subfigure}
    \begin{subfigure}{0.15\textwidth}
\includegraphics[width=\textwidth]{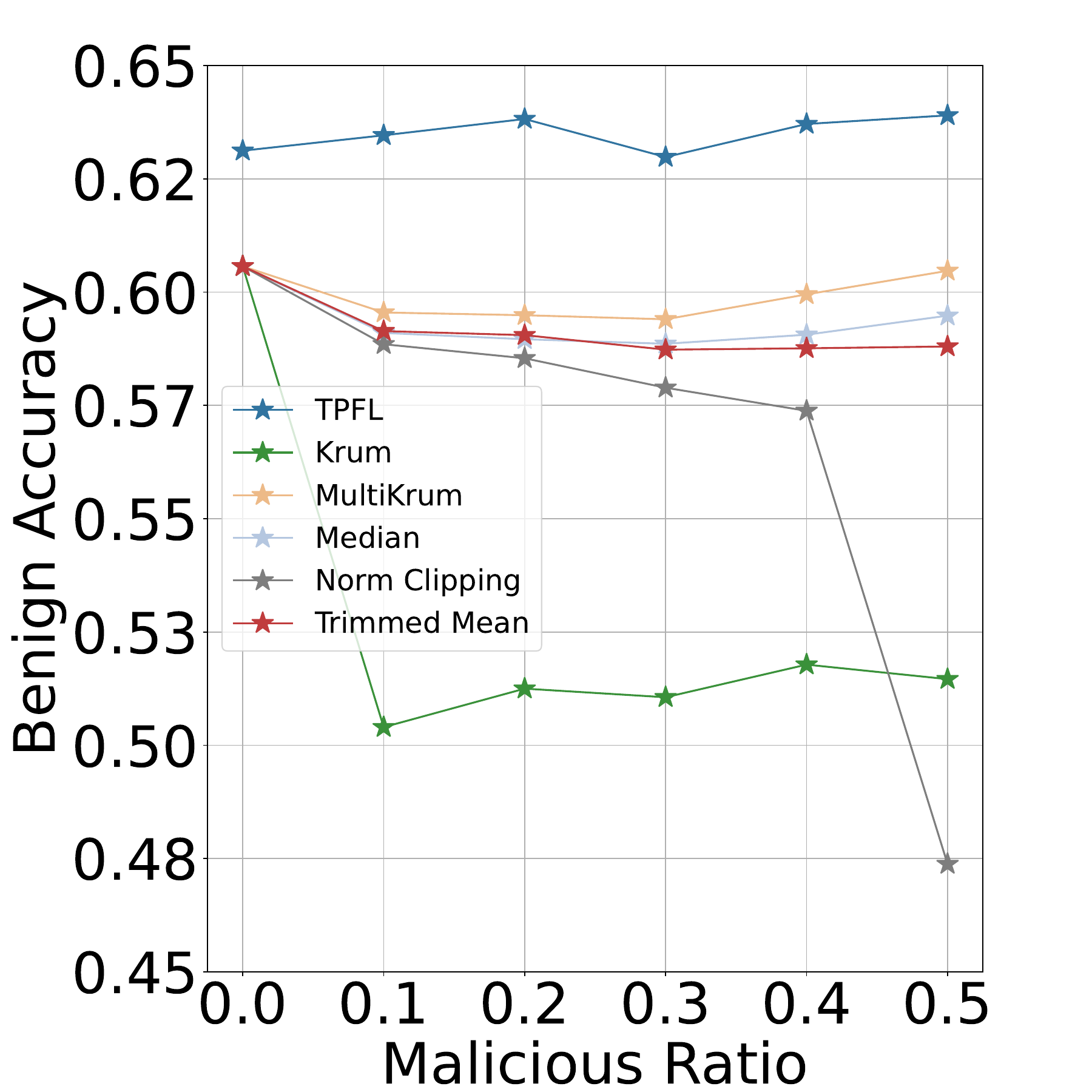}
          \caption{Random \cite{Random}}
          \label{sb}
    \end{subfigure}
    \begin{subfigure}{0.15\textwidth}
\includegraphics[width=\textwidth]{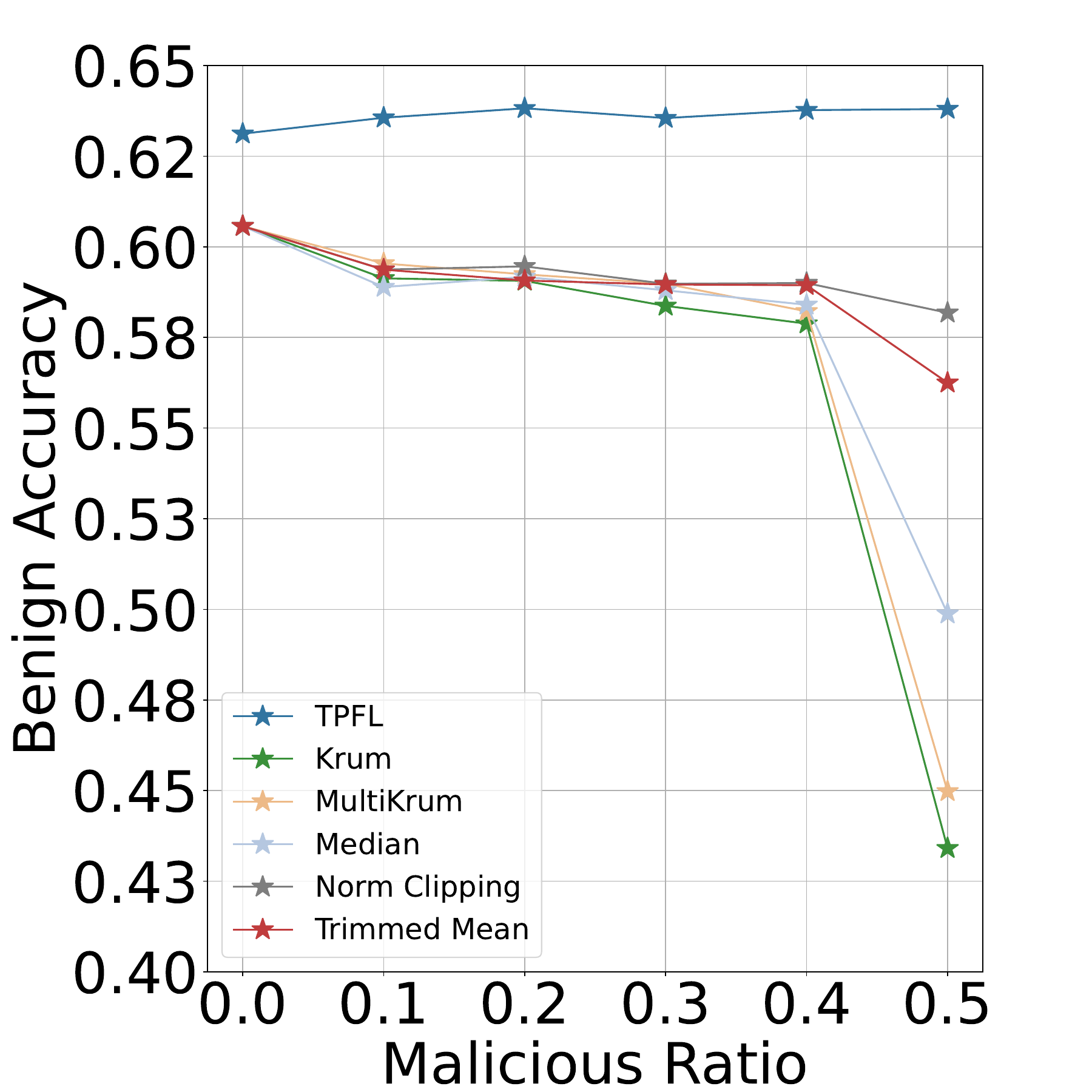}
          \caption{LIE \cite{LIE}}
          \label{sc}
    \end{subfigure}
    \begin{subfigure}{0.15\textwidth}
\includegraphics[width=\textwidth]{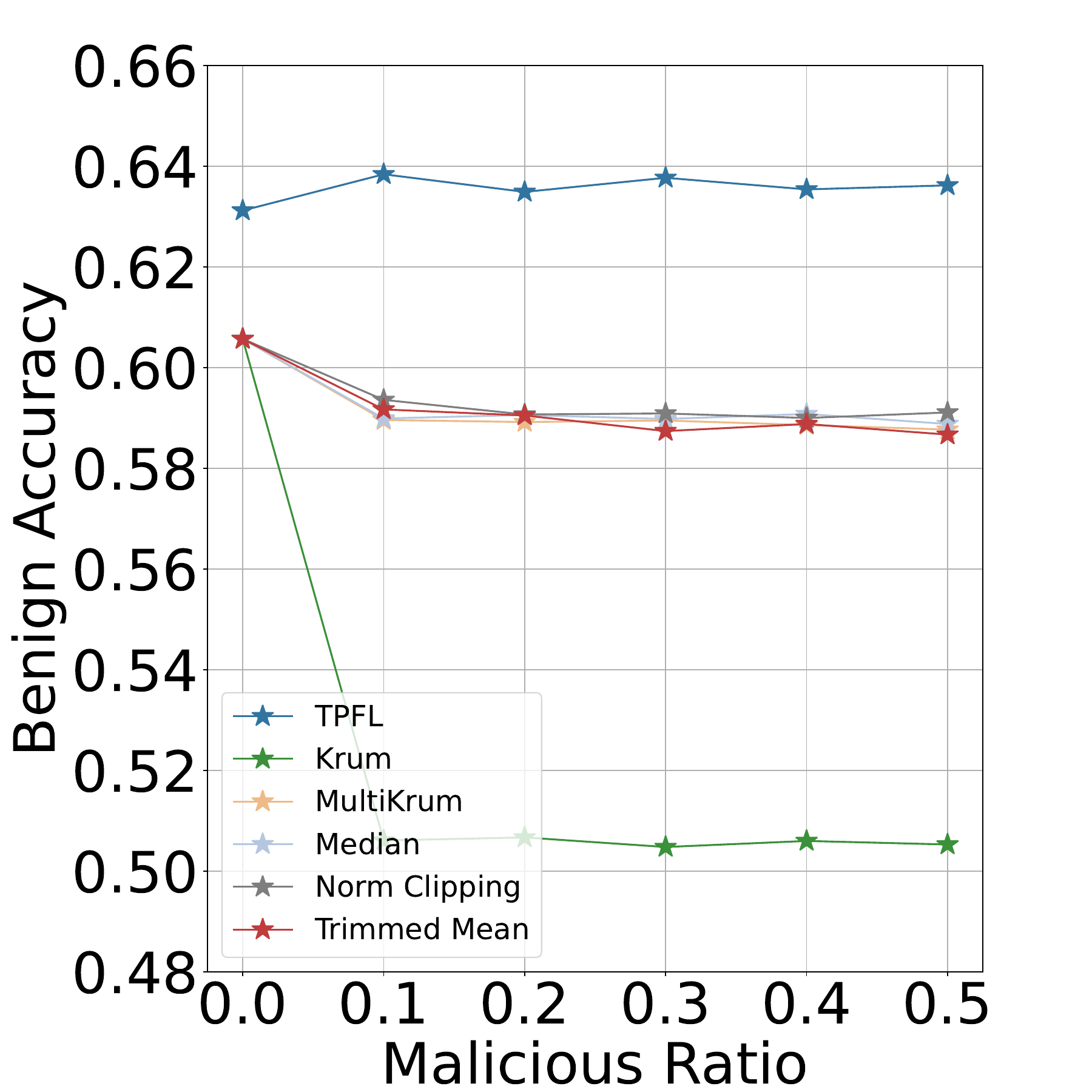}
          \caption{MPAF \cite{MPAF}}
          \label{sd}
    \end{subfigure}
        \begin{subfigure}{0.15\textwidth}
\includegraphics[width=\textwidth]{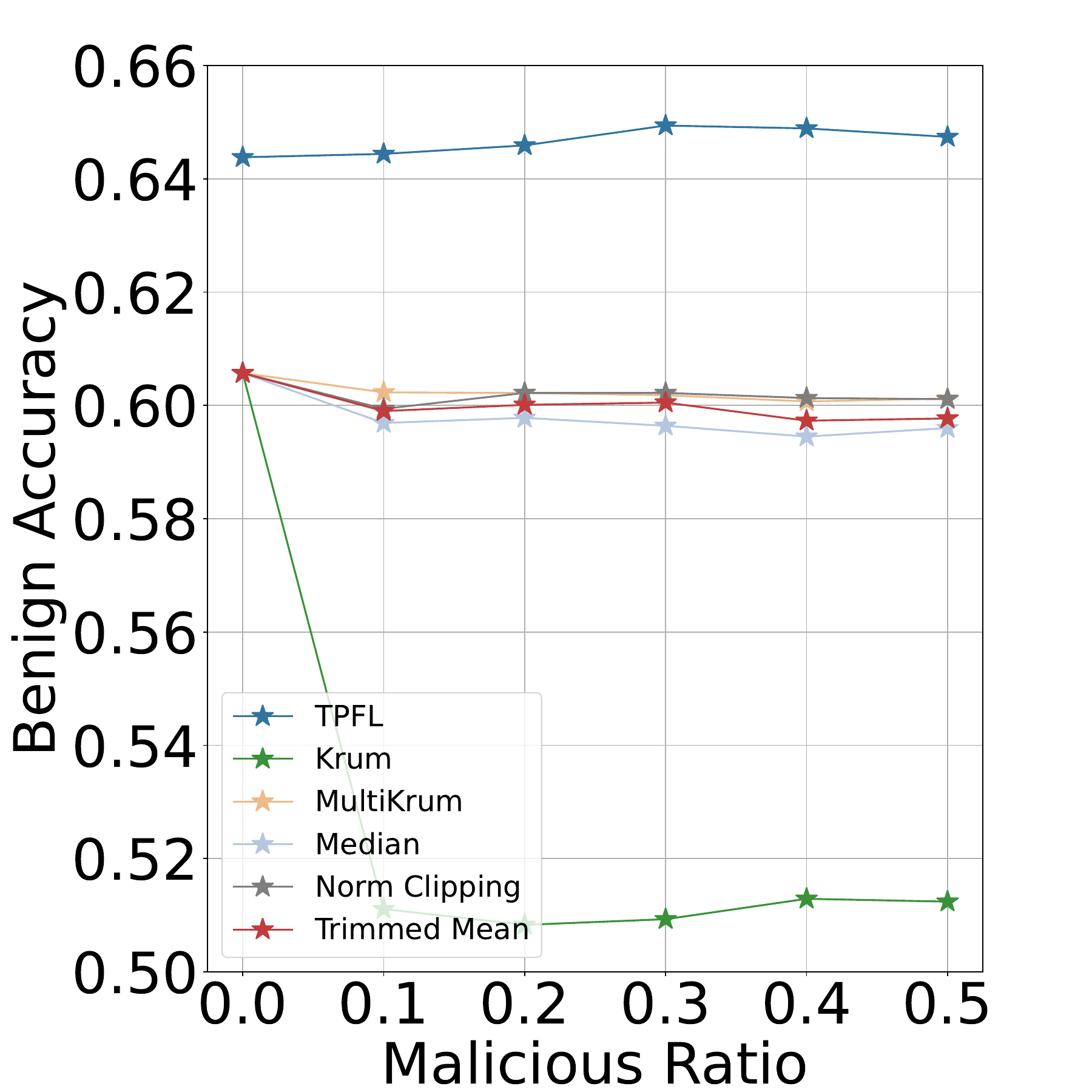}
          \caption{STAT-OPT \cite{LocalModelPoisoning}}
          \label{se}
    \end{subfigure}
    \caption{Safety performance of TPFL. We evaluate the performance under different attack strategies with varied malicious client ratios from 0.1 to 0.5 and compare the performance with SOTA defense methods. As can be seen, TPFL exhibits great resistance against SOTA attacking strategies, showcasing similar safety performance with advanced defense methods.}
    \label{safety}
\end{figure*}

\subsection{Reliability}
We compared the reliability of federated models trained through TPFL with other SOTA federated methods, including FedFOMO \cite{FedFOMO}, FedALA \cite{FedALA}, and federated uncertainty estimation method APH \cite{APH}. Following \citep{EDL, APH}, we evaluate the reliability from both in-distribution and OOD scenarios. For in-domain scenarios, we first evaluate the predictive uncertainty performance of TPFL. Following \citep{EDL}, we utilize uncertainty as a threshold to filter predictions and reject to make decisions on these uncertain instances. As can be seen in Fig.\ref{ra} and \ref{rb}, with the decrease of the uncertainty threshold, the predictive accuracy of TPFL models consistently increases, validating the reliability of instance uncertainty to assist judgments.

We further validate the reliability of TPFL in OOD scenarios. We expect models to reject predicting OOD samples. Specifically, we treat SVHN \cite{SVHN} as the OOD dataset for Cifar10 and Cifar100 \cite{Cifar}. However, as most existing methods can not directly output uncertainty, we utilize predictive entropy \cite{EDL, APH} as uncertainty to conduct a fair comparison. As shown in Fig.\ref{rc} and Fig.\ref{rd}, the reliability of TPFL significantly surpasses the other compared methods, showcasing the higher uncertainty on OOD samples.

\subsection{Security}
We further conduct experiments to validate the security of TPFL. For attacking methods, we utilize Random \cite{Random}, Flip \cite{FlipAttack}, LIE \cite{LIE}, MPAF \cite{MPAF}, STAT-OPT \cite{LocalModelPoisoning} to launch attacks. Median \cite{MedianMeanTrMedian}, Norm-Clipping \cite{NormClip}, Krum \cite{Krum}, and Multi-Krum \cite{Krum} are leveraged to defend those attacks for comparison. We evaluate the performance under those attacks with varied malicious client ratios from 0.1 to 0.5. All experiments are conducted on heterogeneous Cifar10 with 100 clients. Except for TPFL, other defense methods are applied to FedAvg using the same model-split personalized strategy to conduct training. As demonstrated in Fig.\ref{safety}, TPFL exhibits excellent resistance against SOTA attacking strategies, showcasing similar safety performance with advanced defense methods. Note that TPFL achieves higher accuracy without attacks, so it (the blue line) has a different starting point from others.

\begin{table}[t!]
  \centering
  \caption{Training overhead of TPFL compared with other methods in milliseconds. Simulations are conducted in heterogeneous Cifar10 with 10 clients on a single RTX4090 GPU.}
  \resizebox{0.72\linewidth}{!}{
    \begin{tabular}{cccc}
    \toprule
    Method & CNN   & ResNet18 & ResNet50 \\
    \midrule
    FedAvg \cite{FedAvg}& 1154.2±22.1 & 4117.9±40.2 & 8325.4±16.7 \\
    FedProx \cite{FedProx} & 1179.2±6.5 & 4933.9±21.4 & 10387.7±8.1 \\
    FedRoD \cite{FedRoD} & 1240.2±6.4 & 4368.6±96.7 & 8455.4±112.2 \\
    Ditto \cite{ditto} & 1398.8±5.2 & 4709.2±143.5 & 9436.9±84.2 \\
    FedALA \cite{FedALA} & 1525.9±8.8 & 5408.4±126.3 & 11501.1±266.4 \\
    TPFL  & 2129.1±40.8 & 4782.5±202.8 & 9913.9±16.4 \\
    \bottomrule
    \end{tabular}%
    }
  \label{trainingcost}%
\end{table}%

\begin{table}[t!]
  \centering
  \caption{Complexity and time overhead of defense strategies in milliseconds. Simulations are conducted in heterogeneous Cifar10 with a three-layer CNN on a single RTX4090 GPU.}
  \resizebox{0.77\linewidth}{!}{
    \begin{tabular}{ccccc}
    \toprule
    Method & Client 10 & Client 20 & Client 100 & Complexity \\
    \midrule
    Krum \cite{Krum} & 42.5±8.3 & 141.7±5.1 & 3657.1±25.5 & $O(dN^2)$\\
    MultiKrum \cite{Krum}& 47.3±10.1 & 152.5±6.9 & 3642.9±1.2 & $O(dN^2)$ \\
    Median \cite{MedianMeanTrMedian}& 5.8±1.2 & 13.8±4.0 & 35.9±10.2 & $O(dN\operatorname{log}N)$\\
    Trimmed Mean \cite{TrimmedMean} & 6.2±2.5 & 15.4±6.2 & 41.8±12.2 & $O(dN\operatorname{log}N)$ \\
    Norm Clipping \cite{NormClip} & 9.1±1.7 & 14.8±2.6 & 63.5±5.9 & $O(d)$ \\
    TPFL  & 19.7±1.1 & 39.8±8.2 & 204.2±20.4 & $O(HNd)$ \\
    \bottomrule
    \end{tabular}%
    }
  \label{defenseTime}%
\end{table}%

\subsection{Additional Cost}
In addition to performance, computational and communication costs are vital factors in resource-limited federated scenarios. For computational cost, the primary concern of TPFL comes from three perspectives: (1) local training cost, (2) aggregation cost, and (3) inference cost. For a $k$-classification problem in a local client with $n$ samples, the overall computational complexity of TPFL's loss function is $O(kn)$, equal to the FedAvg (Detailed analysis is in Appendix). We also provide the numerical results in Table.\ref{trainingcost} to demonstrate its effective local training. For inference, although inference integrates opinions from three different sources, the computational cost is much less than three times but close to two times the original forward cost, as the feature of the generic encoder can be reused in forwarding in different re-balanced heads. Such a cost is common in practice \cite{FedFusion, FedRoD, DBE}, and also acceptable to achieve reliable and precise judgments.


\begin{table}[t!]
  \centering
  \caption{Ablation study on loss terms in TPFL. Uniform Prior means replacing the heterogeneity prior with a uniform prior. Experiments are conducted in Non-IID scenarios with $\beta=0.1$. }
  \resizebox{\linewidth}{!}{
    \begin{tabular}{ccccccc}
    \toprule
          & TPFL  & No $\mathcal{L}_{\text{cor}}$ & No $\mathcal{L}_{\text{inc}}$ & No $\mathcal{L}_{\text{evi}}$ & No $\mathcal{L}_{\text{neg}}$ & Uniform Prior \\
    \midrule
    Cifar10 & \textbf{0.901} & 0.894 & 0.895 & \textbackslash{} & \textbackslash{} & 0.879 \\
    Cifar100 & \textbf{0.442} & 0.411 & 0.438 & \textbackslash{} & \textbackslash{} & 0.422 \\
    Tiny-ImageNet & \textbf{0.293} & 0.173 & 0.262 & \textbackslash{} & \textbackslash{} & 0.288 \\
    \bottomrule
    \end{tabular}}
  \label{ablation_tab}%
\end{table}%

\begin{figure}[t!]
    \centering
    \begin{subfigure}{0.16\textwidth}
        \includegraphics[width=\textwidth]{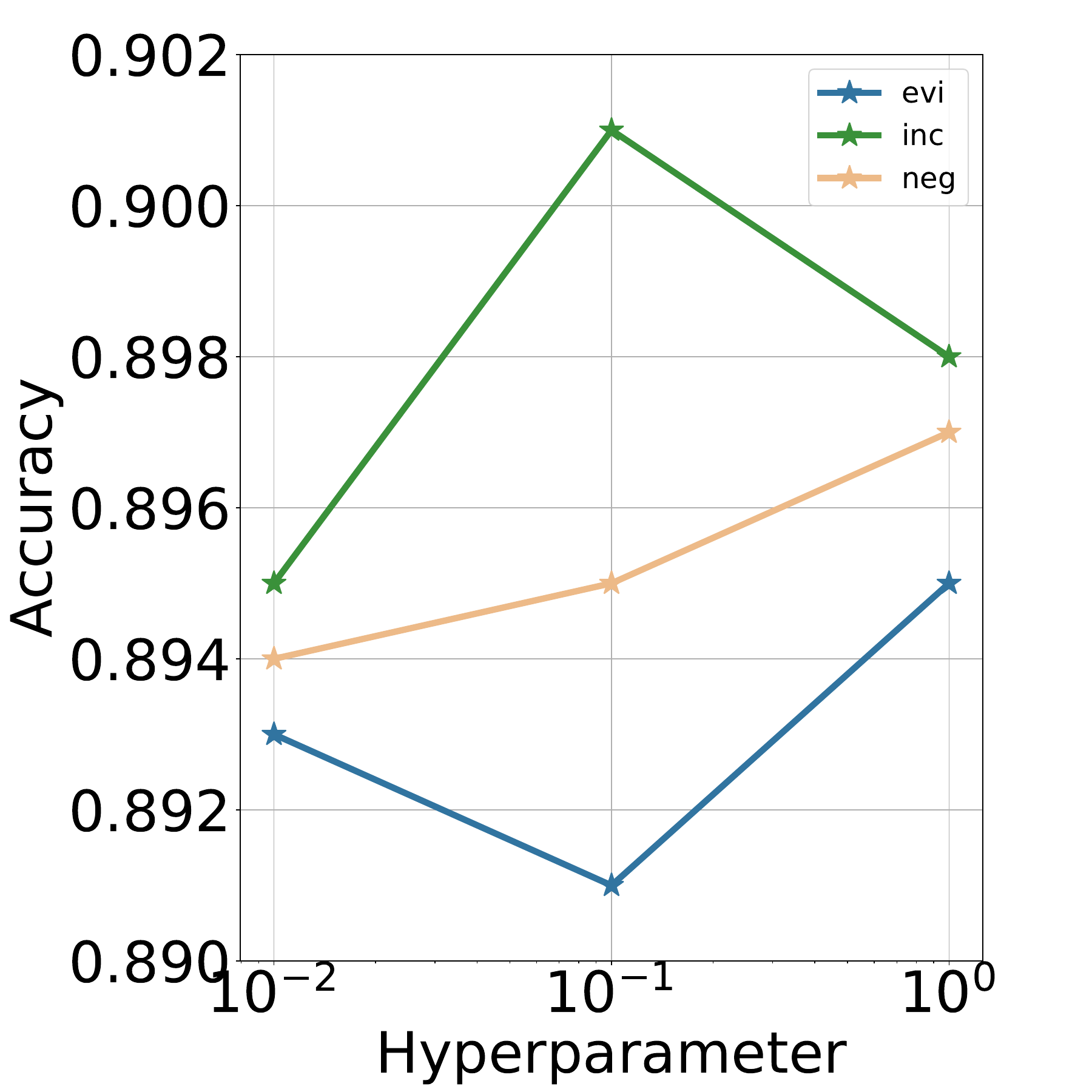}
        \caption{Cifar10}
    \end{subfigure}
    \begin{subfigure}{0.16\textwidth}
        \includegraphics[width=\textwidth]{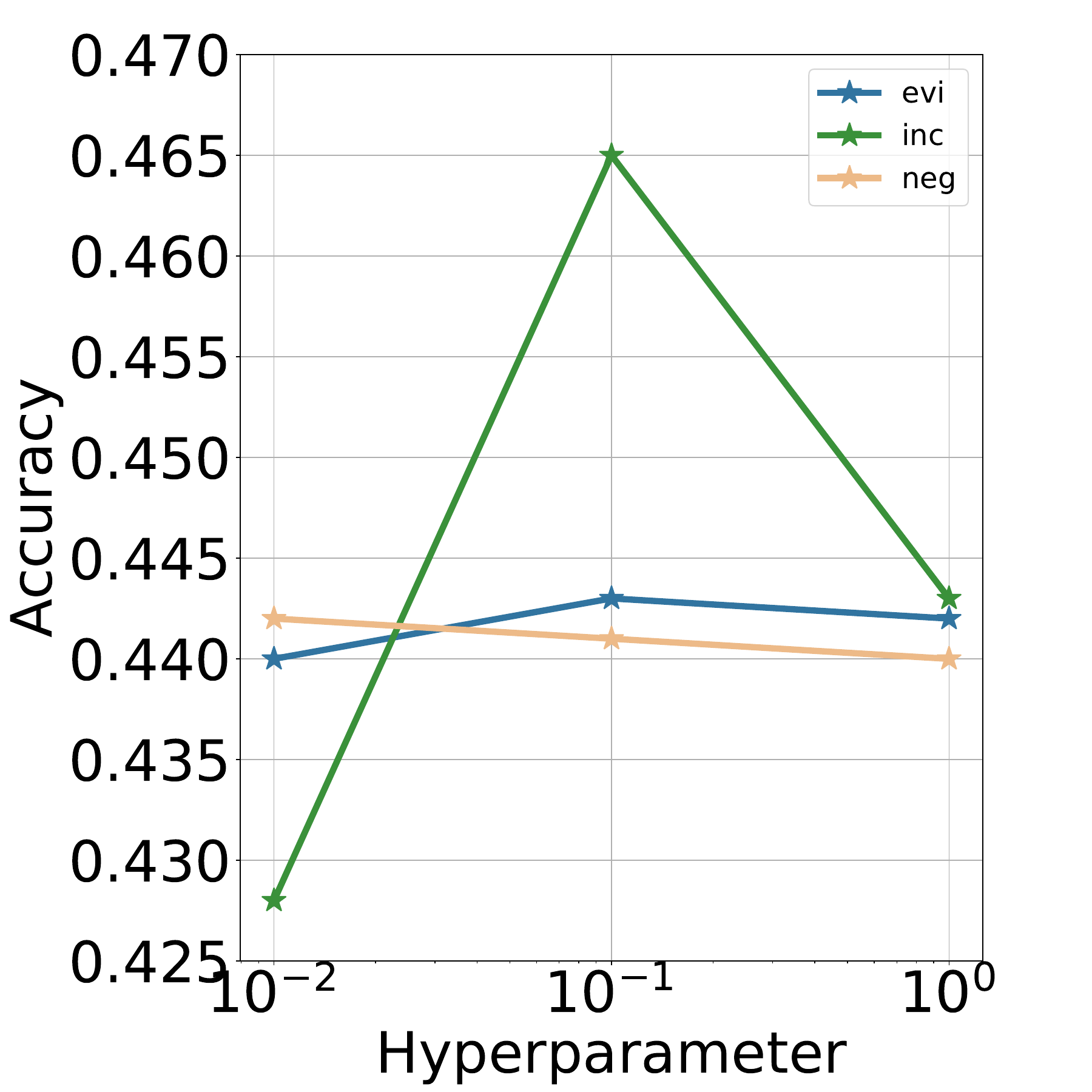}
        \caption{Cifar100}
    \end{subfigure}
    \caption{Ablation study on the sensitivity of hyper-parameters. As can be seen, TPFL is robust to the choice of $\lambda_2$ and $\lambda_3$, while $\lambda_1$ needs a proper choice to balance suppressing fake evidence and encouraging true evidence (e.g., 0.1).}
    \label{fig_param}
\end{figure}

For the cost of model uncertainty estimation, we utilize a holdout set with $H$ samples. In our setting, we set $H=100$, which is trivial for a central server with abundant resources. Assuming the model parameter number is $d$, and the client number is $N$, we report the theoretical analysis of its computation cost with numerical results in Table. \ref{defenseTime}. Detailed analysis can be found in the Appendix. For a small $H$, $O(HNd)$, the aggregation complexity in TPFL,  is close to even smaller than the Median and Trimmed mean, much lower than the complexity of Krum and Multi-Krum in massive-client scenarios. Note that TPFL does not require additional communication compared with FedAvg. Experiments of communication cost are in the Appendix.

\subsection{Ablation Study}
Here, we discuss the design of the loss function and its related hyper-parameters. The loss of TPFL involves five terms, which seems redundant and hard to balance. However, in practice, the $\mathcal{L}_{\text{evi}}$ and $\mathcal{L}_{\text{neg}}$ are proposed to adapt the subjective training to imbalance and skewed data distribution. We report that the pioneer RIPFL \cite{RIPFL}, which applies DST to federated framework directly, can not converge and consistently encounter numerical instability in Non-IID federated scenarios. Except for $\mathcal{L}_{\text{CE}}$, the other two terms are utilized to regularize the evidence. We report the experimental results of TPFL's variants without a particular design in Table.\ref{ablation_tab}. To further discuss the sensitivity of TPFL to its related hyper-parameters $\lambda_1, \lambda_2$ and $\lambda_3$, we conduct experiments on heterogeneous Cifar10 and Cifar100 to explore their impacts. Experimental results are illustrated in Fig.\ref{fig_param}. As demonstrated, TPFL is robust to the choice of $\lambda_2$ and $\lambda_3$ (Corresponded to $\mathcal{L}_{\text{evi}}$ and $\mathcal{L}_{\text{neg}}$), while $\lambda_1$ which controls the weight of $\mathcal{L}_{\text{inc}}$ need a good choice to balance suppressing fake evidence and encouraging true evidence. We suggest $\lambda_1$ to be set to $0.1$ for all experiments.

\section{Conclusion}
In this paper, we extend the subjective logic to the federated scenarios and establish Trustworthy Personalized Federated Learning (TPFL). TPFL enables trustworthy inference combined with heterogeneity prior to handling data heterogeneity. TPFL can also directly estimate the instance uncertainty and model uncertainty, which are further utilized to ensure the reliability and safety of TPFL. Our experiments validate that TPFL achieves performance, reliability, and safety. 
{
    \small
    \bibliographystyle{ieeenat_fullname}
    \bibliography{main}

\begin{thebibliography}{97}
\providecommand{\natexlab}[1]{#1}
\providecommand{\url}[1]{\texttt{#1}}
\expandafter\ifx\csname urlstyle\endcsname\relax
  \providecommand{\doi}[1]{doi: #1}\else
  \providecommand{\doi}{doi: \begingroup \urlstyle{rm}\Url}\fi

\bibitem[Acar et~al.(2021)Acar, Zhao, Navarro, Mattina, Whatmough, and Saligrama]{FedDyn}
Durmus Alp~Emre Acar, Yue Zhao, Ramon~Matas Navarro, Matthew Mattina, Paul~N Whatmough, and Venkatesh Saligrama.
\newblock Federated learning based on dynamic regularization.
\newblock \emph{arXiv preprint arXiv:2111.04263}, 2021.

\bibitem[Alazab et~al.(2021{\natexlab{a}})Alazab, RM, Parimala, Maddikunta, Gadekallu, and Pham]{FLSurvey}
Mamoun Alazab, Swarna~Priya RM, M Parimala, Praveen Kumar~Reddy Maddikunta, Thippa~Reddy Gadekallu, and Quoc-Viet Pham.
\newblock Federated learning for cybersecurity: Concepts, challenges, and future directions.
\newblock \emph{IEEE Transactions on Industrial Informatics}, 18\penalty0 (5):\penalty0 3501--3509, 2021{\natexlab{a}}.

\bibitem[Alazab et~al.(2021{\natexlab{b}})Alazab, RM, Parimala, Maddikunta, Gadekallu, and Pham]{FLSurvey2}
Mamoun Alazab, Swarna~Priya RM, M Parimala, Praveen Kumar~Reddy Maddikunta, Thippa~Reddy Gadekallu, and Quoc-Viet Pham.
\newblock Federated learning for cybersecurity: Concepts, challenges, and future directions.
\newblock \emph{IEEE Transactions on Industrial Informatics}, 18\penalty0 (5):\penalty0 3501--3509, 2021{\natexlab{b}}.

\bibitem[Bagdasaryan et~al.(2019)Bagdasaryan, Poursaeed, and Shmatikov]{DPHarm}
Eugene Bagdasaryan, Omid Poursaeed, and Vitaly Shmatikov.
\newblock Differential privacy has disparate impact on model accuracy.
\newblock \emph{Advances in neural information processing systems}, 32, 2019.

\bibitem[Bagdasaryan et~al.(2020{\natexlab{a}})Bagdasaryan, Veit, Hua, Estrin, and Shmatikov]{BackdoorFL}
Eugene Bagdasaryan, Andreas Veit, Yiqing Hua, Deborah Estrin, and Vitaly Shmatikov.
\newblock How to backdoor federated learning.
\newblock In \emph{International conference on artificial intelligence and statistics}, pages 2938--2948. PMLR, 2020{\natexlab{a}}.

\bibitem[Bagdasaryan et~al.(2020{\natexlab{b}})Bagdasaryan, Veit, Hua, Estrin, and Shmatikov]{ModelReplacement}
Eugene Bagdasaryan, Andreas Veit, Yiqing Hua, Deborah Estrin, and Vitaly Shmatikov.
\newblock How to backdoor federated learning.
\newblock In \emph{International conference on artificial intelligence and statistics}, pages 2938--2948. PMLR, 2020{\natexlab{b}}.

\bibitem[Baruch et~al.(2019)Baruch, Baruch, and Goldberg]{LIE}
Gilad Baruch, Moran Baruch, and Yoav Goldberg.
\newblock A little is enough: Circumventing defenses for distributed learning.
\newblock \emph{Advances in Neural Information Processing Systems}, 32, 2019.

\bibitem[Biggio et~al.(2012)Biggio, Nelson, and Laskov]{FlipAttack}
Battista Biggio, Blaine Nelson, and Pavel Laskov.
\newblock Poisoning attacks against support vector machines.
\newblock \emph{arXiv preprint arXiv:1206.6389}, 2012.

\bibitem[Blanchard et~al.(2017)Blanchard, El~Mhamdi, Guerraoui, and Stainer]{Krum}
Peva Blanchard, El~Mahdi El~Mhamdi, Rachid Guerraoui, and Julien Stainer.
\newblock Machine learning with adversaries: Byzantine tolerant gradient descent.
\newblock \emph{Advances in neural information processing systems}, 30, 2017.

\bibitem[Briggs et~al.(2020)Briggs, Fan, and Andras]{HRCFL}
Christopher Briggs, Zhong Fan, and Peter Andras.
\newblock Federated learning with hierarchical clustering of local updates to improve training on non-iid data.
\newblock In \emph{2020 international joint conference on neural networks (IJCNN)}, pages 1--9. IEEE, 2020.

\bibitem[Byrd and Polychroniadou(2020)]{FLFinance}
David Byrd and Antigoni Polychroniadou.
\newblock Differentially private secure multi-party computation for federated learning in financial applications.
\newblock In \emph{Proceedings of the First ACM International Conference on AI in Finance}, pages 1--9, 2020.

\bibitem[Caesar et~al.(2020)Caesar, Bankiti, Lang, Vora, Liong, Xu, Krishnan, Pan, Baldan, and Beijbom]{drive}
Holger Caesar, Varun Bankiti, Alex~H Lang, Sourabh Vora, Venice~Erin Liong, Qiang Xu, Anush Krishnan, Yu Pan, Giancarlo Baldan, and Oscar Beijbom.
\newblock nuscenes: A multimodal dataset for autonomous driving.
\newblock In \emph{Proceedings of the IEEE/CVF conference on computer vision and pattern recognition}, pages 11621--11631, 2020.

\bibitem[Cao et~al.(2022)Cao, Wang, Chen, Jiang, Zhang, Tian, and Wang]{medical}
Hu Cao, Yueyue Wang, Joy Chen, Dongsheng Jiang, Xiaopeng Zhang, Qi Tian, and Manning Wang.
\newblock Swin-unet: Unet-like pure transformer for medical image segmentation.
\newblock In \emph{European conference on computer vision}, pages 205--218. Springer, 2022.

\bibitem[Cao and Gong(2022)]{MPAF}
Xiaoyu Cao and Neil~Zhenqiang Gong.
\newblock Mpaf: Model poisoning attacks to federated learning based on fake clients.
\newblock In \emph{Proceedings of the IEEE/CVF Conference on Computer Vision and Pattern Recognition}, pages 3396--3404, 2022.

\bibitem[Castelvecchi(2016)]{blackbox1}
Davide Castelvecchi.
\newblock Can we open the black box of ai?
\newblock \emph{Nature News}, 538\penalty0 (7623):\penalty0 20, 2016.

\bibitem[Chakraborty et~al.(2018)Chakraborty, Alam, Dey, Chattopadhyay, and Mukhopadhyay]{adver1}
Anirban Chakraborty, Manaar Alam, Vishal Dey, Anupam Chattopadhyay, and Debdeep Mukhopadhyay.
\newblock Adversarial attacks and defences: A survey.
\newblock \emph{arXiv preprint arXiv:1810.00069}, 2018.

\bibitem[Chen and Chao(2022)]{FedRoD}
Hong-You Chen and Wei-Lun Chao.
\newblock On bridging generic and personalized federated learning for image classification.
\newblock In \emph{International Conference on Learning Representations}, 2022.

\bibitem[Chen et~al.(2023)Chen, Zhu, and Zheng]{FedKA}
Jinqian Chen, Jihua Zhu, and Qinghai Zheng.
\newblock Towards fast and stable federated learning: Confronting heterogeneity via knowledge anchor.
\newblock In \emph{Proceedings of the 31st ACM International Conference on Multimedia}, pages 8697--8706, 2023.

\bibitem[Chen et~al.(2024)Chen, Zhu, Zheng, Li, and Tian]{APH}
Jinqian Chen, Jihua Zhu, Qinghai Zheng, Zhongyu Li, and Zhiqiang Tian.
\newblock Watch your head: Assembling projection heads to save the reliability of federated models.
\newblock \emph{Proceedings of the AAAI Conference on Artificial Intelligence}, 2024.

\bibitem[Chen et~al.(2022)Chen, Lou, He, Bai, and Deng]{chen2022evidential}
Liang Chen, Yihang Lou, Jianzhong He, Tao Bai, and Minghua Deng.
\newblock Evidential neighborhood contrastive learning for universal domain adaptation.
\newblock In \emph{Proceedings of the AAAI Conference on Artificial Intelligence}, pages 6258--6267, 2022.

\bibitem[Chen et~al.(2017)Chen, Liu, Li, Lu, and Song]{TargetedBackdoor}
Xinyun Chen, Chang Liu, Bo Li, Kimberly Lu, and Dawn Song.
\newblock Targeted backdoor attacks on deep learning systems using data poisoning.
\newblock \emph{arXiv preprint arXiv:1712.05526}, 2017.

\bibitem[Collins et~al.(2021)Collins, Hassani, Mokhtari, and Shakkottai]{FedRep}
Liam Collins, Hamed Hassani, Aryan Mokhtari, and Sanjay Shakkottai.
\newblock Exploiting shared representations for personalized federated learning.
\newblock In \emph{International conference on machine learning}, pages 2089--2099. PMLR, 2021.

\bibitem[Dayan et~al.(2021)Dayan, Roth, Zhong, Harouni, Gentili, Abidin, Liu, Costa, Wood, Tsai, et~al.]{FLMedical}
Ittai Dayan, Holger~R Roth, Aoxiao Zhong, Ahmed Harouni, Amilcare Gentili, Anas~Z Abidin, Andrew Liu, Anthony~Beardsworth Costa, Bradford~J Wood, Chien-Sung Tsai, et~al.
\newblock Federated learning for predicting clinical outcomes in patients with covid-19.
\newblock \emph{Nature medicine}, 27\penalty0 (10):\penalty0 1735--1743, 2021.

\bibitem[Dempster(1967)]{DST}
AP Dempster.
\newblock Upper and lower probabilities induced by a multivalued mapping.
\newblock \emph{The Annals of Mathematical Statistics}, 38\penalty0 (2):\penalty0 325--339, 1967.

\bibitem[Fallah et~al.(2020)Fallah, Mokhtari, and Ozdaglar]{Per-FedAvg}
Alireza Fallah, Aryan Mokhtari, and Asuman Ozdaglar.
\newblock Personalized federated learning with theoretical guarantees: A model-agnostic meta-learning approach.
\newblock \emph{Advances in Neural Information Processing Systems}, 33:\penalty0 3557--3568, 2020.

\bibitem[Fang et~al.(2020)Fang, Cao, Jia, and Gong]{LocalModelPoisoning}
Minghong Fang, Xiaoyu Cao, Jinyuan Jia, and Neil Gong.
\newblock Local model poisoning attacks to $\{$Byzantine-Robust$\}$ federated learning.
\newblock In \emph{29th USENIX security symposium (USENIX Security 20)}, pages 1605--1622, 2020.

\bibitem[Fukushima(1975)]{ReLU}
Kunihiko Fukushima.
\newblock Cognitron: A self-organizing multilayered neural network.
\newblock \emph{Biological cybernetics}, 20\penalty0 (3):\penalty0 121--136, 1975.

\bibitem[Gal and Ghahramani(2016)]{Dropout}
Yarin Gal and Zoubin Ghahramani.
\newblock Dropout as a bayesian approximation: Representing model uncertainty in deep learning.
\newblock In \emph{international conference on machine learning}, pages 1050--1059. PMLR, 2016.

\bibitem[Gao et~al.(2022)Gao, Fu, Li, Chen, Xu, and Xu]{FedDC}
Liang Gao, Huazhu Fu, Li Li, Yingwen Chen, Ming Xu, and Cheng-Zhong Xu.
\newblock Feddc: Federated learning with non-iid data via local drift decoupling and correction.
\newblock In \emph{Proceedings of the IEEE/CVF conference on computer vision and pattern recognition}, pages 10112--10121, 2022.

\bibitem[Gawlikowski et~al.(2021)Gawlikowski, Tassi, Ali, Lee, Humt, Feng, Kruspe, Triebel, Jung, Roscher, et~al.]{UncertaintySurvey}
Jakob Gawlikowski, Cedrique Rovile~Njieutcheu Tassi, Mohsin Ali, Jongseok Lee, Matthias Humt, Jianxiang Feng, Anna Kruspe, Rudolph Triebel, Peter Jung, Ribana Roscher, et~al.
\newblock A survey of uncertainty in deep neural networks.
\newblock \emph{arXiv preprint arXiv:2107.03342}, 2021.

\bibitem[Gawlikowski et~al.(2023)Gawlikowski, Tassi, Ali, Lee, Humt, Feng, Kruspe, Triebel, Jung, Roscher, et~al.]{UncertaintySurvey1}
Jakob Gawlikowski, Cedrique Rovile~Njieutcheu Tassi, Mohsin Ali, Jongseok Lee, Matthias Humt, Jianxiang Feng, Anna Kruspe, Rudolph Triebel, Peter Jung, Ribana Roscher, et~al.
\newblock A survey of uncertainty in deep neural networks.
\newblock \emph{Artificial Intelligence Review}, 56\penalty0 (Suppl 1):\penalty0 1513--1589, 2023.

\bibitem[Ghosh et~al.(2019)Ghosh, Hong, Yin, and Ramchandran]{RobustFLCluster}
Avishek Ghosh, Justin Hong, Dong Yin, and Kannan Ramchandran.
\newblock Robust federated learning in a heterogeneous environment.
\newblock \emph{arXiv preprint arXiv:1906.06629}, 2019.

\bibitem[Guo et~al.(2017)Guo, Pleiss, Sun, and Weinberger]{ECE}
Chuan Guo, Geoff Pleiss, Yu Sun, and Kilian~Q Weinberger.
\newblock On calibration of modern neural networks.
\newblock In \emph{International conference on machine learning}, pages 1321--1330. PMLR, 2017.

\bibitem[Han et~al.(2022)Han, Zhang, Fu, and Zhou]{TMC}
Zongbo Han, Changqing Zhang, Huazhu Fu, and Joey~Tianyi Zhou.
\newblock Trusted multi-view classification with dynamic evidential fusion.
\newblock \emph{IEEE transactions on pattern analysis and machine intelligence}, 45\penalty0 (2):\penalty0 2551--2566, 2022.

\bibitem[Hsu et~al.(2019)Hsu, Qi, and Brown]{FedAvgM}
Tzu-Ming~Harry Hsu, Hang Qi, and Matthew Brown.
\newblock Measuring the effects of non-identical data distribution for federated visual classification.
\newblock \emph{arXiv preprint arXiv:1909.06335}, 2019.

\bibitem[Huang et~al.(2021)Huang, Chu, Zhou, Wang, Liu, Pei, and Zhang]{FedAMP}
Yutao Huang, Lingyang Chu, Zirui Zhou, Lanjun Wang, Jiangchuan Liu, Jian Pei, and Yong Zhang.
\newblock Personalized cross-silo federated learning on non-iid data.
\newblock In \emph{Proceedings of the AAAI conference on artificial intelligence}, pages 7865--7873, 2021.

\bibitem[Isaak and Hanna(2018)]{dataprivacy}
Jim Isaak and Mina~J Hanna.
\newblock User data privacy: Facebook, cambridge analytica, and privacy protection.
\newblock \emph{Computer}, 51\penalty0 (8):\penalty0 56--59, 2018.

\bibitem[J{\o}sang(1997)]{SLReasoning}
Audun J{\o}sang.
\newblock Artificial reasoning with subjective logic.
\newblock In \emph{Proceedings of the second Australian workshop on commonsense reasoning}, page~34. Citeseer, 1997.

\bibitem[J{\o}sang(2016)]{SubjectiveLogic}
Audun J{\o}sang.
\newblock \emph{Subjective logic}.
\newblock Springer, 2016.

\bibitem[J{\o}sang and Hankin(2012)]{SLrule1}
Audun J{\o}sang and Robin Hankin.
\newblock Interpretation and fusion of hyper opinions in subjective logic.
\newblock In \emph{2012 15th International Conference on Information Fusion}, pages 1225--1232. IEEE, 2012.

\bibitem[Karimireddy et~al.(2020)Karimireddy, Kale, Mohri, Reddi, Stich, and Suresh]{SCAFFOLD}
Sai~Praneeth Karimireddy, Satyen Kale, Mehryar Mohri, Sashank Reddi, Sebastian Stich, and Ananda~Theertha Suresh.
\newblock Scaffold: Stochastic controlled averaging for federated learning.
\newblock In \emph{International conference on machine learning}, pages 5132--5143. PMLR, 2020.

\bibitem[Krizhevsky et~al.(2009)Krizhevsky, Hinton, et~al.]{Cifar}
Alex Krizhevsky, Geoffrey Hinton, et~al.
\newblock Learning multiple layers of features from tiny images.
\newblock 2009.

\bibitem[Lakshminarayanan et~al.(2017)Lakshminarayanan, Pritzel, and Blundell]{DeepEnsemble}
Balaji Lakshminarayanan, Alexander Pritzel, and Charles Blundell.
\newblock Simple and scalable predictive uncertainty estimation using deep ensembles.
\newblock \emph{Advances in neural information processing systems}, 30, 2017.

\bibitem[Lee et~al.(2022)Lee, Jeong, Shin, Bae, and Yun]{FedNTD}
Gihun Lee, Minchan Jeong, Yongjin Shin, Sangmin Bae, and Se-Young Yun.
\newblock Preservation of the global knowledge by not-true distillation in federated learning.
\newblock \emph{Advances in Neural Information Processing Systems}, 35:\penalty0 38461--38474, 2022.

\bibitem[Li et~al.(2023{\natexlab{a}})Li, Qi, Liu, Di, Liu, Pei, Yi, and Zhou]{trustworthyAISurvey2}
Bo Li, Peng Qi, Bo Liu, Shuai Di, Jingen Liu, Jiquan Pei, Jinfeng Yi, and Bowen Zhou.
\newblock Trustworthy ai: From principles to practices.
\newblock \emph{ACM Computing Surveys}, 55\penalty0 (9):\penalty0 1--46, 2023{\natexlab{a}}.

\bibitem[Li et~al.(2021{\natexlab{a}})Li, He, and Song]{MOON}
Qinbin Li, Bingsheng He, and Dawn Song.
\newblock Model-contrastive federated learning.
\newblock In \emph{Proceedings of the IEEE/CVF conference on computer vision and pattern recognition}, pages 10713--10722, 2021{\natexlab{a}}.

\bibitem[Li et~al.(2022{\natexlab{a}})Li, Diao, Chen, and He]{Exp_NonIID}
Qinbin Li, Yiqun Diao, Quan Chen, and Bingsheng He.
\newblock Federated learning on non-iid data silos: An experimental study.
\newblock In \emph{2022 IEEE 38th International Conference on Data Engineering (ICDE)}, pages 965--978. IEEE, 2022{\natexlab{a}}.

\bibitem[Li et~al.(2022{\natexlab{b}})Li, Diao, Chen, and He]{ExperimentalStudy}
Qinbin Li, Yiqun Diao, Quan Chen, and Bingsheng He.
\newblock Federated learning on non-iid data silos: An experimental study.
\newblock In \emph{2022 IEEE 38th International Conference on Data Engineering (ICDE)}, pages 965--978. IEEE, 2022{\natexlab{b}}.

\bibitem[Li et~al.(2020)Li, Sahu, Zaheer, Sanjabi, Talwalkar, and Smith]{FedProx}
Tian Li, Anit~Kumar Sahu, Manzil Zaheer, Maziar Sanjabi, Ameet Talwalkar, and Virginia Smith.
\newblock Federated optimization in heterogeneous networks.
\newblock \emph{Proceedings of Machine learning and systems}, 2:\penalty0 429--450, 2020.

\bibitem[Li et~al.(2021{\natexlab{b}})Li, Hu, Beirami, and Smith]{ditto}
Tian Li, Shengyuan Hu, Ahmad Beirami, and Virginia Smith.
\newblock Ditto: Fair and robust federated learning through personalization.
\newblock In \emph{International Conference on Machine Learning}, pages 6357--6368. PMLR, 2021{\natexlab{b}}.

\bibitem[Li et~al.(2023{\natexlab{b}})Li, Shang, He, Lin, and Wu]{FedETF}
Zexi Li, Xinyi Shang, Rui He, Tao Lin, and Chao Wu.
\newblock No fear of classifier biases: Neural collapse inspired federated learning with synthetic and fixed classifier.
\newblock In \emph{Proceedings of the IEEE/CVF International Conference on Computer Vision}, pages 5319--5329, 2023{\natexlab{b}}.

\bibitem[Lin et~al.(2020)Lin, Kong, Stich, and Jaggi]{FedDF}
Tao Lin, Lingjing Kong, Sebastian~U Stich, and Martin Jaggi.
\newblock Ensemble distillation for robust model fusion in federated learning.
\newblock \emph{Advances in Neural Information Processing Systems}, 33:\penalty0 2351--2363, 2020.

\bibitem[Linsner et~al.(2021)Linsner, Adilova, D{\"a}ubener, Kamp, and Fischer]{PioneerUncertaintyFL}
Florian Linsner, Linara Adilova, Sina D{\"a}ubener, Michael Kamp, and Asja Fischer.
\newblock Approaches to uncertainty quantification in federated deep learning.
\newblock In \emph{Joint European Conference on Machine Learning and Knowledge Discovery in Databases}, pages 128--145. Springer, 2021.

\bibitem[Liu et~al.(2022)Liu, Wang, Fan, Liu, Li, Jain, Liu, Jain, and Tang]{trustworthyAIsurvey1}
Haochen Liu, Yiqi Wang, Wenqi Fan, Xiaorui Liu, Yaxin Li, Shaili Jain, Yunhao Liu, Anil Jain, and Jiliang Tang.
\newblock Trustworthy ai: A computational perspective.
\newblock \emph{ACM Transactions on Intelligent Systems and Technology}, 14\penalty0 (1):\penalty0 1--59, 2022.

\bibitem[Lu et~al.(2023)Lu, Yu, Karimireddy, Jordan, and Raskar]{FCP}
Charles Lu, Yaodong Yu, Sai~Praneeth Karimireddy, Michael Jordan, and Ramesh Raskar.
\newblock Federated conformal predictors for distributed uncertainty quantification.
\newblock In \emph{International Conference on Machine Learning}, pages 22942--22964. PMLR, 2023.

\bibitem[Luo and Wu(2022)]{APPLE}
Jun Luo and Shandong Wu.
\newblock Adapt to adaptation: Learning personalization for cross-silo federated learning.
\newblock In \emph{IJCAI: proceedings of the conference}, page 2166. NIH Public Access, 2022.

\bibitem[Luo et~al.(2023)Luo, Li, Lan, and Gao]{GradMA}
Kangyang Luo, Xiang Li, Yunshi Lan, and Ming Gao.
\newblock Gradma: A gradient-memory-based accelerated federated learning with alleviated catastrophic forgetting.
\newblock In \emph{Proceedings of the IEEE/CVF Conference on Computer Vision and Pattern Recognition}, pages 3708--3717, 2023.

\bibitem[Lyu et~al.(2020)Lyu, Yu, and Yang]{FLThreatSurvey}
Lingjuan Lyu, Han Yu, and Qiang Yang.
\newblock Threats to federated learning: A survey.
\newblock \emph{arXiv preprint arXiv:2003.02133}, 2020.

\bibitem[McMahan et~al.(2017)McMahan, Moore, Ramage, Hampson, and y~Arcas]{FedAvg}
Brendan McMahan, Eider Moore, Daniel Ramage, Seth Hampson, and Blaise~Aguera y Arcas.
\newblock Communication-efficient learning of deep networks from decentralized data.
\newblock In \emph{Artificial intelligence and statistics}, pages 1273--1282. PMLR, 2017.

\bibitem[Mills et~al.(2021)Mills, Hu, and Min]{pFedMTL}
Jed Mills, Jia Hu, and Geyong Min.
\newblock Multi-task federated learning for personalised deep neural networks in edge computing.
\newblock \emph{IEEE Transactions on Parallel and Distributed Systems}, 33\penalty0 (3):\penalty0 630--641, 2021.

\bibitem[Netzer et~al.(2011)Netzer, Wang, Coates, Bissacco, Wu, Ng, et~al.]{SVHN}
Yuval Netzer, Tao Wang, Adam Coates, Alessandro Bissacco, Baolin Wu, Andrew~Y Ng, et~al.
\newblock Reading digits in natural images with unsupervised feature learning.
\newblock In \emph{NIPS workshop on deep learning and unsupervised feature learning}, page~4. Granada, 2011.

\bibitem[Nguyen et~al.(2021)Nguyen, Ding, Pathirana, Seneviratne, Li, and Poor]{FLCity}
Dinh~C Nguyen, Ming Ding, Pubudu~N Pathirana, Aruna Seneviratne, Jun Li, and H~Vincent Poor.
\newblock Federated learning for internet of things: A comprehensive survey.
\newblock \emph{IEEE Communications Surveys \& Tutorials}, 23\penalty0 (3):\penalty0 1622--1658, 2021.

\bibitem[Ozdayi et~al.(2021)Ozdayi, Kantarcioglu, and Gel]{RobustLR}
Mustafa~Safa Ozdayi, Murat Kantarcioglu, and Yulia~R Gel.
\newblock Defending against backdoors in federated learning with robust learning rate.
\newblock In \emph{Proceedings of the AAAI Conference on Artificial Intelligence}, pages 9268--9276, 2021.

\bibitem[Pandey and Yu(2023)]{Red}
Deep~Shankar Pandey and Qi Yu.
\newblock Learn to accumulate evidence from all training samples: Theory and practice.
\newblock In \emph{International Conference on Machine Learning}, pages 26963--26989. PMLR, 2023.

\bibitem[Pillutla et~al.(2022{\natexlab{a}})Pillutla, Kakade, and Harchaoui]{RobustAggregation}
Krishna Pillutla, Sham~M Kakade, and Zaid Harchaoui.
\newblock Robust aggregation for federated learning.
\newblock \emph{IEEE Transactions on Signal Processing}, 70:\penalty0 1142--1154, 2022{\natexlab{a}}.

\bibitem[Pillutla et~al.(2022{\natexlab{b}})Pillutla, Malik, Mohamed, Rabbat, Sanjabi, and Xiao]{FedAlt}
Krishna Pillutla, Kshitiz Malik, Abdel-Rahman Mohamed, Mike Rabbat, Maziar Sanjabi, and Lin Xiao.
\newblock Federated learning with partial model personalization.
\newblock In \emph{International Conference on Machine Learning}, pages 17716--17758. PMLR, 2022{\natexlab{b}}.

\bibitem[Qin et~al.(2023)Qin, Yang, Wang, Han, and Hu]{RIPFL}
Zixuan Qin, Liu Yang, Qilong Wang, Yahong Han, and Qinghua Hu.
\newblock Reliable and interpretable personalized federated learning.
\newblock In \emph{Proceedings of the IEEE/CVF Conference on Computer Vision and Pattern Recognition}, pages 20422--20431, 2023.

\bibitem[Rodr{\'\i}guez-Barroso et~al.(2023)Rodr{\'\i}guez-Barroso, Jim{\'e}nez-L{\'o}pez, Luz{\'o}n, Herrera, and Mart{\'\i}nez-C{\'a}mara]{ThreatSurvey}
Nuria Rodr{\'\i}guez-Barroso, Daniel Jim{\'e}nez-L{\'o}pez, M~Victoria Luz{\'o}n, Francisco Herrera, and Eugenio Mart{\'\i}nez-C{\'a}mara.
\newblock Survey on federated learning threats: Concepts, taxonomy on attacks and defences, experimental study and challenges.
\newblock \emph{Information Fusion}, 90:\penalty0 148--173, 2023.

\bibitem[Sattler et~al.(2020{\natexlab{a}})Sattler, M{\"u}ller, Wiegand, and Samek]{ClusterFL}
Felix Sattler, Klaus-Robert M{\"u}ller, Thomas Wiegand, and Wojciech Samek.
\newblock On the byzantine robustness of clustered federated learning.
\newblock In \emph{ICASSP 2020-2020 IEEE International Conference on Acoustics, Speech and Signal Processing (ICASSP)}, pages 8861--8865. IEEE, 2020{\natexlab{a}}.

\bibitem[Sattler et~al.(2020{\natexlab{b}})Sattler, M{\"u}ller, Wiegand, and Samek]{RobustCFL}
Felix Sattler, Klaus-Robert M{\"u}ller, Thomas Wiegand, and Wojciech Samek.
\newblock On the byzantine robustness of clustered federated learning.
\newblock In \emph{ICASSP 2020-2020 IEEE International Conference on Acoustics, Speech and Signal Processing (ICASSP)}, pages 8861--8865. IEEE, 2020{\natexlab{b}}.

\bibitem[Sensoy et~al.(2018)Sensoy, Kaplan, and Kandemir]{EDL}
Murat Sensoy, Lance Kaplan, and Melih Kandemir.
\newblock Evidential deep learning to quantify classification uncertainty.
\newblock \emph{Advances in neural information processing systems}, 31, 2018.

\bibitem[Sentz and Ferson(2002)]{SLrule0}
Kari Sentz and Scott Ferson.
\newblock Combination of evidence in dempster-shafer theory.
\newblock 2002.

\bibitem[Shafer(1976)]{DSTBook}
Glenn Shafer.
\newblock \emph{A mathematical theory of evidence}.
\newblock Princeton university press, 1976.

\bibitem[Shayan et~al.(2020)Shayan, Fung, Yoon, and Beschastnikh]{Biscotti}
Muhammad Shayan, Clement Fung, Chris~JM Yoon, and Ivan Beschastnikh.
\newblock Biscotti: A blockchain system for private and secure federated learning.
\newblock \emph{IEEE Transactions on Parallel and Distributed Systems}, 32\penalty0 (7):\penalty0 1513--1525, 2020.

\bibitem[Shejwalkar et~al.(2022)Shejwalkar, Houmansadr, Kairouz, and Ramage]{PoisonEvaluation}
Virat Shejwalkar, Amir Houmansadr, Peter Kairouz, and Daniel Ramage.
\newblock Back to the drawing board: A critical evaluation of poisoning attacks on production federated learning.
\newblock In \emph{2022 IEEE Symposium on Security and Privacy (SP)}, pages 1354--1371. IEEE, 2022.

\bibitem[Shen et~al.(2016)Shen, Tople, and Saxena]{AUROR}
Shiqi Shen, Shruti Tople, and Prateek Saxena.
\newblock Auror: Defending against poisoning attacks in collaborative deep learning systems.
\newblock In \emph{Proceedings of the 32nd Annual Conference on Computer Security Applications}, pages 508--519, 2016.

\bibitem[Shi et~al.(2023)Shi, Lai, Al~Kontar, and Chowdhury]{FedEnsemble}
Naichen Shi, Fan Lai, Raed Al~Kontar, and Mosharaf Chowdhury.
\newblock Ensemble models in federated learning for improved generalization and uncertainty quantification.
\newblock \emph{IEEE Transactions on Automation Science and Engineering}, 2023.

\bibitem[Short et~al.(2020)Short, Leligou, Papoutsidakis, and Theocharis]{BlockChainSecurity}
Andrew~Ronald Short, Helen~C Leligou, Michael Papoutsidakis, and Efstathios Theocharis.
\newblock Using blockchain technologies to improve security in federated learning systems.
\newblock In \emph{2020 IEEE 44th annual computers, software, and applications conference (COMPSAC)}, pages 1183--1188. IEEE, 2020.

\bibitem[Shwartz-Ziv and Tishby(2017)]{blackbox2}
Ravid Shwartz-Ziv and Naftali Tishby.
\newblock Opening the black box of deep neural networks via information.
\newblock \emph{arXiv preprint arXiv:1703.00810}, 2017.

\bibitem[Smith et~al.(2017{\natexlab{a}})Smith, Chiang, Sanjabi, and Talwalkar]{FedMTL}
Virginia Smith, Chao-Kai Chiang, Maziar Sanjabi, and Ameet~S Talwalkar.
\newblock Federated multi-task learning.
\newblock \emph{Advances in neural information processing systems}, 30, 2017{\natexlab{a}}.

\bibitem[Smith et~al.(2017{\natexlab{b}})Smith, Chiang, Sanjabi, and Talwalkar]{MOCHA}
Virginia Smith, Chao-Kai Chiang, Maziar Sanjabi, and Ameet~S Talwalkar.
\newblock Federated multi-task learning.
\newblock \emph{Advances in neural information processing systems}, 30, 2017{\natexlab{b}}.

\bibitem[Sun et~al.(2022)Sun, Dou, Yang, Zhang, Wang, Philip, He, and Li]{adver2}
Lichao Sun, Yingtong Dou, Carl Yang, Kai Zhang, Ji Wang, S~Yu Philip, Lifang He, and Bo Li.
\newblock Adversarial attack and defense on graph data: A survey.
\newblock \emph{IEEE Transactions on Knowledge and Data Engineering}, 2022.

\bibitem[Sun et~al.(2019)Sun, Kairouz, Suresh, and McMahan]{NormClip}
Ziteng Sun, Peter Kairouz, Ananda~Theertha Suresh, and H~Brendan McMahan.
\newblock Can you really backdoor federated learning?
\newblock \emph{arXiv preprint arXiv:1911.07963}, 2019.

\bibitem[T~Dinh et~al.(2020)T~Dinh, Tran, and Nguyen]{pFedMe}
Canh T~Dinh, Nguyen Tran, and Josh Nguyen.
\newblock Personalized federated learning with moreau envelopes.
\newblock \emph{Advances in Neural Information Processing Systems}, 33:\penalty0 21394--21405, 2020.

\bibitem[Tan et~al.(2022)Tan, Long, Ma, Liu, Zhou, and Jiang]{FedPCL}
Yue Tan, Guodong Long, Jie Ma, Lu Liu, Tianyi Zhou, and Jing Jiang.
\newblock Federated learning from pre-trained models: A contrastive learning approach.
\newblock \emph{Advances in Neural Information Processing Systems}, 35:\penalty0 19332--19344, 2022.

\bibitem[Wang et~al.(2020)Wang, Sreenivasan, Rajput, Vishwakarma, Agarwal, Sohn, Lee, and Papailiopoulos]{AttackTail}
Hongyi Wang, Kartik Sreenivasan, Shashank Rajput, Harit Vishwakarma, Saurabh Agarwal, Jy-yong Sohn, Kangwook Lee, and Dimitris Papailiopoulos.
\newblock Attack of the tails: Yes, you really can backdoor federated learning.
\newblock \emph{Advances in Neural Information Processing Systems}, 33:\penalty0 16070--16084, 2020.

\bibitem[Xu and Lyu(2020)]{Random}
Xinyi Xu and Lingjuan Lyu.
\newblock A reputation mechanism is all you need: Collaborative fairness and adversarial robustness in federated learning.
\newblock \emph{arXiv preprint arXiv:2011.10464}, 2020.

\bibitem[Yao et~al.(2019)Yao, Huang, Wu, Zhang, and Sun]{FedFusion}
Xin Yao, Tianchi Huang, Chenglei Wu, Rui-Xiao Zhang, and Lifeng Sun.
\newblock Towards faster and better federated learning: A feature fusion approach.
\newblock In \emph{2019 IEEE International Conference on Image Processing (ICIP)}, pages 175--179. IEEE, 2019.

\bibitem[Yin et~al.(2018{\natexlab{a}})Yin, Chen, Kannan, and Bartlett]{MedianMeanTrMedian}
Dong Yin, Yudong Chen, Ramchandran Kannan, and Peter Bartlett.
\newblock Byzantine-robust distributed learning: Towards optimal statistical rates.
\newblock In \emph{International conference on machine learning}, pages 5650--5659. Pmlr, 2018{\natexlab{a}}.

\bibitem[Yin et~al.(2018{\natexlab{b}})Yin, Chen, Kannan, and Bartlett]{TrimmedMean}
Dong Yin, Yudong Chen, Ramchandran Kannan, and Peter Bartlett.
\newblock Byzantine-robust distributed learning: Towards optimal statistical rates.
\newblock In \emph{International Conference on Machine Learning}, pages 5650--5659. PMLR, 2018{\natexlab{b}}.

\bibitem[Zhang et~al.(2023{\natexlab{a}})Zhang, Hua, Wang, Song, Xue, Ma, Cao, and Guan]{GPFL}
Jianqing Zhang, Yang Hua, Hao Wang, Tao Song, Zhengui Xue, Ruhui Ma, Jian Cao, and Haibing Guan.
\newblock Gpfl: Simultaneously learning global and personalized feature information for personalized federated learning.
\newblock In \emph{Proceedings of the IEEE/CVF International Conference on Computer Vision}, pages 5041--5051, 2023{\natexlab{a}}.

\bibitem[Zhang et~al.(2023{\natexlab{b}})Zhang, Hua, Wang, Song, Xue, Ma, and Guan]{FedALA}
Jianqing Zhang, Yang Hua, Hao Wang, Tao Song, Zhengui Xue, Ruhui Ma, and Haibing Guan.
\newblock Fedala: Adaptive local aggregation for personalized federated learning.
\newblock In \emph{Proceedings of the AAAI Conference on Artificial Intelligence}, pages 11237--11244, 2023{\natexlab{b}}.

\bibitem[Zhang et~al.(2024)Zhang, Hua, Cao, Wang, Song, Xue, Ma, and Guan]{DBE}
Jianqing Zhang, Yang Hua, Jian Cao, Hao Wang, Tao Song, Zhengui Xue, Ruhui Ma, and Haibing Guan.
\newblock Eliminating domain bias for federated learning in representation space.
\newblock \emph{Advances in Neural Information Processing Systems}, 36, 2024.

\bibitem[Zhang et~al.(2021)Zhang, Sapra, Fidler, Yeung, and Alvarez]{FedFOMO}
Michael Zhang, Karan Sapra, Sanja Fidler, Serena Yeung, and Jose~M Alvarez.
\newblock Personalized federated learning with first order model optimization.
\newblock In \emph{International Conference on Learning Representations}, 2021.

\bibitem[Zhang et~al.(2019)Zhang, Yao, Sun, and Tay]{recommend}
Shuai Zhang, Lina Yao, Aixin Sun, and Yi Tay.
\newblock Deep learning based recommender system: A survey and new perspectives.
\newblock \emph{ACM computing surveys (CSUR)}, 52\penalty0 (1):\penalty0 1--38, 2019.

\bibitem[Zhang et~al.(2023{\natexlab{c}})Zhang, Zeng, Luo, Xu, and King]{TFLSurvey}
Yifei Zhang, Dun Zeng, Jinglong Luo, Zenglin Xu, and Irwin King.
\newblock A survey of trustworthy federated learning with perspectives on security, robustness, and privacy.
\newblock \emph{arXiv preprint arXiv:2302.10637}, 2023{\natexlab{c}}.

\bibitem[Zhu et~al.(2021)Zhu, Xu, Liu, and Jin]{FLNonIIDSurvey}
Hangyu Zhu, Jinjin Xu, Shiqing Liu, and Yaochu Jin.
\newblock Federated learning on non-iid data: A survey.
\newblock \emph{Neurocomputing}, 465:\penalty0 371--390, 2021.

\end{thebibliography}
}
\clearpage
\setcounter{page}{1}
\maketitlesupplementary

\section{Detailed Related Work}
\textbf{Federated Learning on Non-IID Data}. Federated learning is a promising training schema for decentralized machine learning since FedAvg \cite{FedAvg} is proposed. However, it struggles in heterogeneous data with performance degradation and convergence retardation \cite{Exp_NonIID}. Numerous adaption algorithms are proposed to overcome this issue, forming generic federated learning (G-FL) methods. G-FL methods aim to train a single generalized global model for all clients and utilize different strategies to combat the heterogeneity, such as client drift mitigation \cite{FedProx, MOON, FedDC}, gradient correction \cite{SCAFFOLD, FedAvgM, FedDyn, GradMA}, and knowledge distillation \cite{FedDF, FedNTD, FedKA}. Unlike G-FL methods, personalized federated learning (P-FL) methods aim to train a personalized model for each participating client to fit their local data. The idea of P-FL methods is first proposed in \citep{FedMTL} and further extended by \citep{MOCHA, pFedMe, Per-FedAvg}. Current SOTA P-FL methods designed for Non-IID data can be roughly divided into following categories according to its personalization strategies: (1) model splitting and parameter decoupling \cite{FedRep, FedRoD}; (2) model mixture \cite{pFedMe, ditto, GPFL}; (3) personalized aggregation \cite{FedFOMO, FedAMP, FedALA}; (4) clustering \cite{ClusterFL, HRCFL}; (5) multitask learning \cite{FedMTL, pFedMTL, GPFL}, and others \cite{FedAlt, FedPCL}. Though achieving superior performance, most current P-FL methods lack the comprehensive fusion of judgments from both generic and personalized models, significantly affecting the precision and reliability of predictive decisions.

\textbf{Safety in Federated Learning.} Safety has been viewed as a critical property of federated learning since its proposal \citep{FedAvg}. It expects federated learning frameworks to have the ability to combat adversarial attacks, ensuring safe, collaborative training and deployment without being backdoored or getting destructive performance degradation by potential malicious clients \cite{ThreatSurvey, FLThreatSurvey}. According to their objective, malicious attacks can be classified into targeted and untargeted attacks. Targeted attacks aim to inject a backdoor (i.e., trigger) to the federated models without harming the overall accuracy so that any instances with the injected trigger will be classified to the desired class \cite{BackdoorFL, TargetedBackdoor}, while untargeted attacks eager to severely degrade the overall performance of the training model \cite{PoisonEvaluation}. This paper mainly focuses on the untargeted attacks that lead to destructive and unacceptable performance across all clients. From the strategies the attacks applied, the untargeted attacks can be further split into data poisoning untargeted attacks \cite{FlipAttack} and model poisoning untargeted attacks \cite{ModelReplacement, LocalModelPoisoning, LIE, PoisonEvaluation, MPAF}. Numerous defense methods have been proposed to combat malicious attacks, which can be classified into robust aggregation-based \cite{MedianMeanTrMedian, RobustAggregation} and anomaly detection-based methods to ensure the safety of federated learning. Aggregation-based methods apply statistically robust operator, such as utilizing median \cite{MedianMeanTrMedian, RobustAggregation, Krum}, trimmed median \cite{MedianMeanTrMedian, RobustAggregation}, and trimmed mean \cite{MedianMeanTrMedian}, or selecting distance-based representative model \cite{Krum}, to replace the traditional simple averaging schema, expecting eliminating the malicious updates via such aggregations. Anomaly detection-based methods \cite{RobustCFL} aims to identify adversarial clients as anomalous data according to the distribution of local updates through clustering \cite{RobustCFL, RobustFLCluster} and block-chain-based filtering \cite{Biscotti, BlockChainSecurity}. Besides these methods, other strategies, such as robust learning rates \cite{RobustLR}, are also applied to combat malicious updates and achieve success. Although great success has been achieved, existing defense methods must make significant trade-offs between performance and safety. Efficient, safe, and highly performed federated learning still needs to be established.

\textbf{Reliability of Federated Learning.} Reliability is an essential aspect of a trustworthy federated learning framework. Here, to clarify the difference between security and reliability of federated learning, we follow the definition of previous work \citep{EDL, APH, ECE}, which expects federated models to provide reliable uncertainty, i.e. to be well-calibrated on in-distribution data and keep sensitive and express high uncertainty when encounters OOD or adversarial data \citep{ UncertaintySurvey1, FedEnsemble, DeepEnsemble}. Recent studies have demonstrated that federated models trained on heterogeneous data are likelier to be unreliable \citep{APH}. To mitigate this issue, numerous works have endeavored to transfer traditional uncertainty estimation methods to federated scenarios \citep{PioneerUncertaintyFL, FedEnsemble}, such as MC-Dropout \citep{Dropout}, Deep Ensembles \citep{DeepEnsemble}, etc, or try to propose novel methods suitable for the distributed training scenarios \citep{FCP, APH}. Although significant progress has been made, most previous work requires additional computation costs, significantly constraining their applications.

\textbf{Subjective Logic (SL)} is a generalized theory of Dempster-Shafer Evidence Theory (DST) \citep{DST, DSTBook}. Unlike classic probability theory, it requires the cooperation of prior knowledge with non-informative weight to assign belief masses to all possible states, naturally introducing the concept of epistemic uncertainty. By collecting evidence from different observations, SL allows evidence from different observation sources to be fused by various combination rules \cite{SLrule0, SLrule1}, updating their prior and belief mass assignments. From bijective mapping between Dirichlet distribution and belief assignments with prior, SL can easily estimate the marginal expectation of the particular state, providing reliable judgments. Recently, EDL \cite{EDL} successfully utilizes the DST, the simplified version of SL, in deep learning fields to quantify model uncertainty, raising its wide application into different practical multi-source scenarios \cite{TMC, Red, chen2022evidential}. Recently, RIPFL \cite{RIPFL} first applies DST into federated scenarios and succeeds wildly in the IID setting. However, the disregard for prior distributions of DST not only leads to the incompleteness of the applied theory \cite{SubjectiveLogic} but is also unacceptable in practical federated scenarios, particularly in Non-IID settings where data distributions are heterogeneous \cite{Exp_NonIID} among all clients.

\begin{figure}[t!]
\captionsetup[subfigure]{justification=centering}
    \centering
      \begin{subfigure}{0.235\textwidth}
        \includegraphics[width=\textwidth]{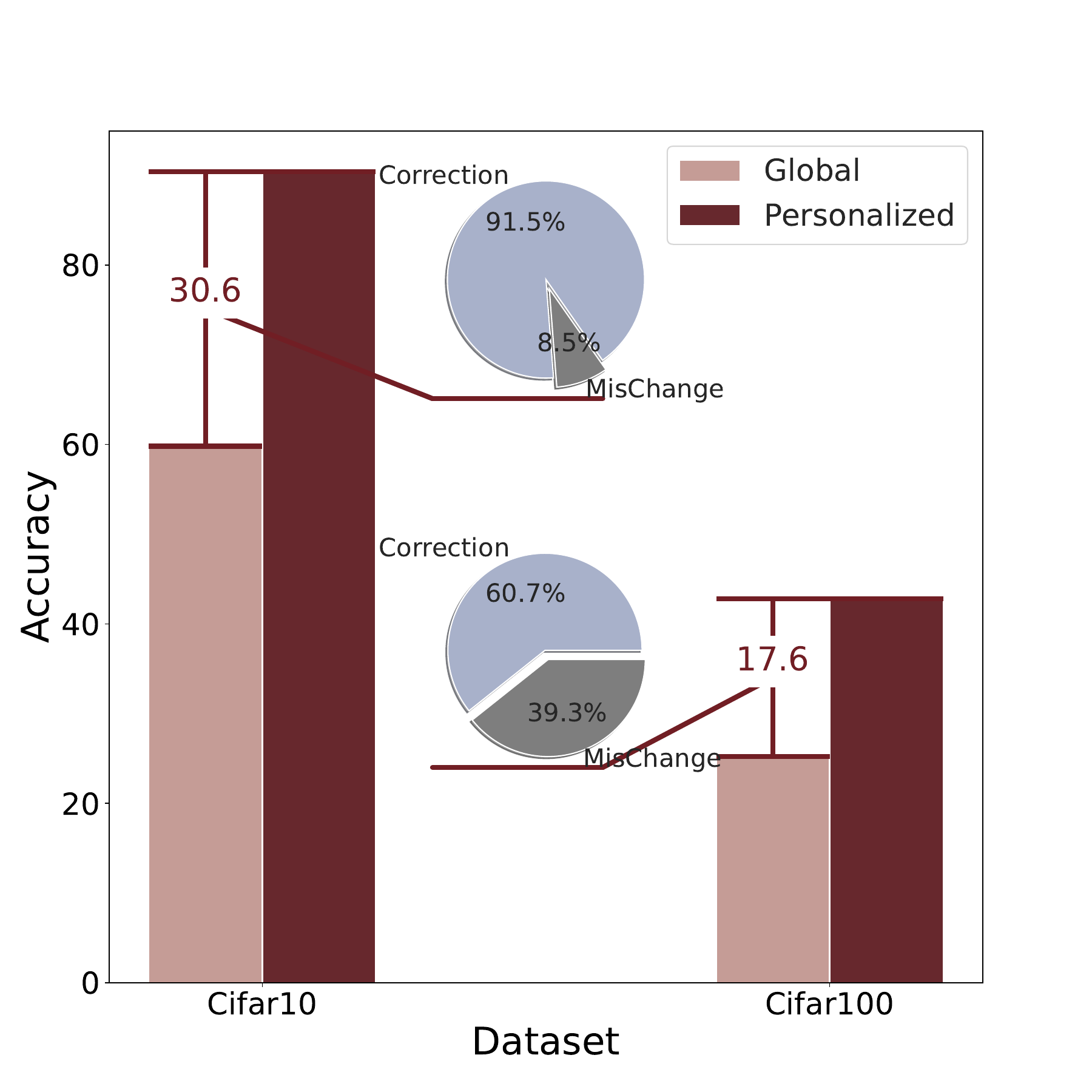}
          \caption{Generic FL    \\(e.g. FedAvg)}
          \label{fig2a}
      \end{subfigure}
      \begin{subfigure}{0.235\textwidth}
        \includegraphics[width=\textwidth]{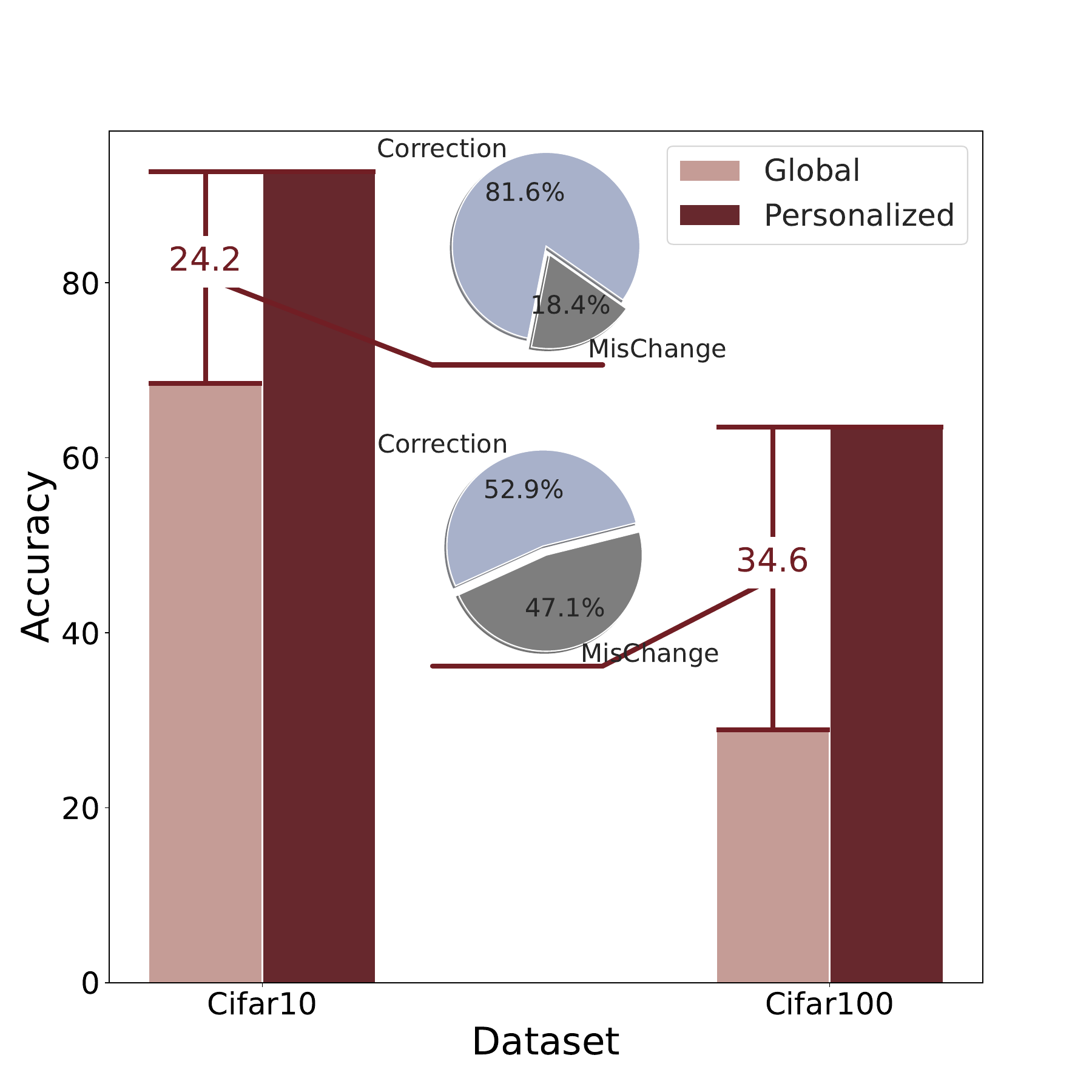}
          \caption{Personalized FL \\(e.g. FedRoD)}
          \label{fig2b}
      \end{subfigure}
            \begin{subfigure}{0.22\textwidth}
        \includegraphics[width=\textwidth]{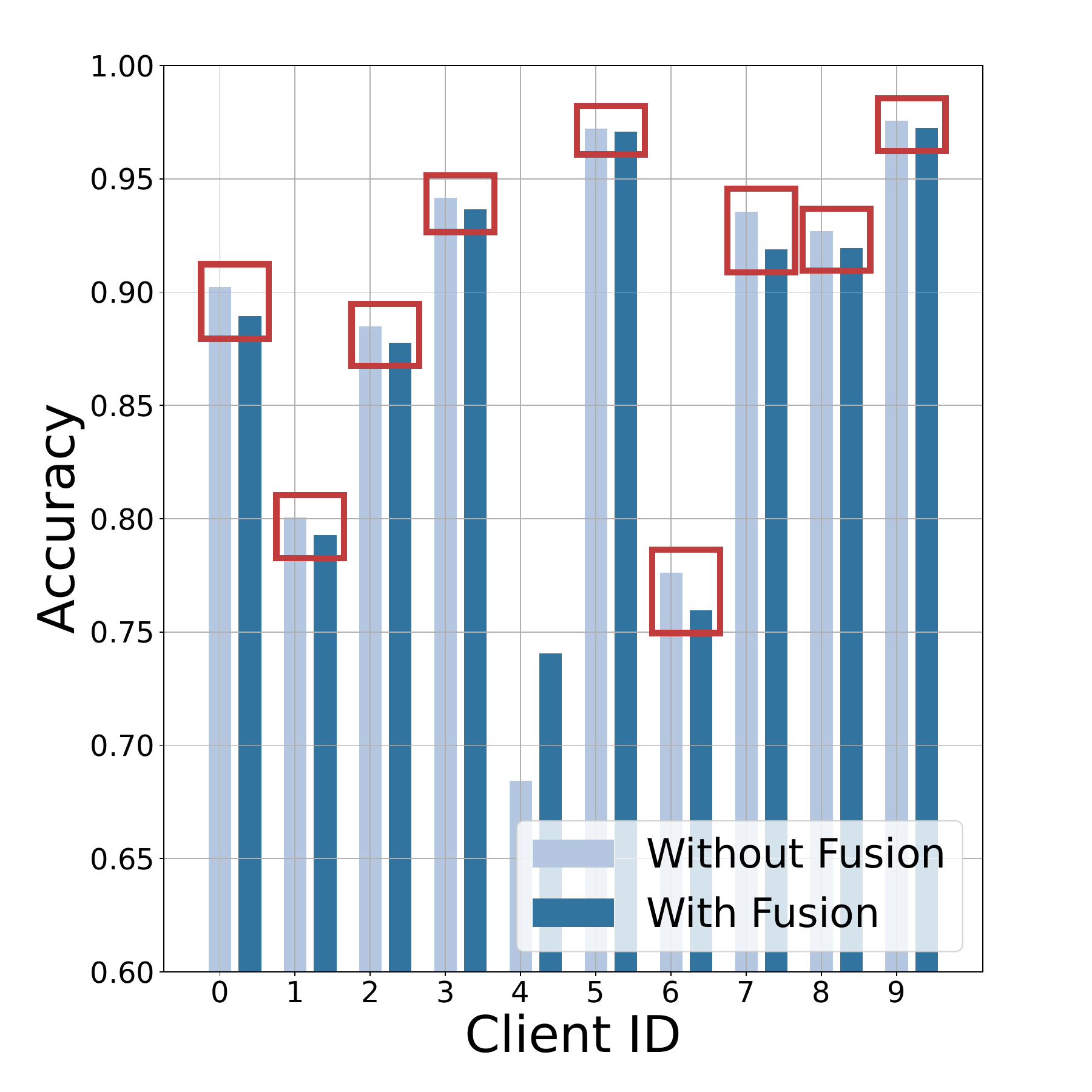}
          \caption{Feature Fusion Client-wise}
          \label{fig3a}
      \end{subfigure}
      \begin{subfigure}{0.22\textwidth}
        \includegraphics[width=\textwidth]{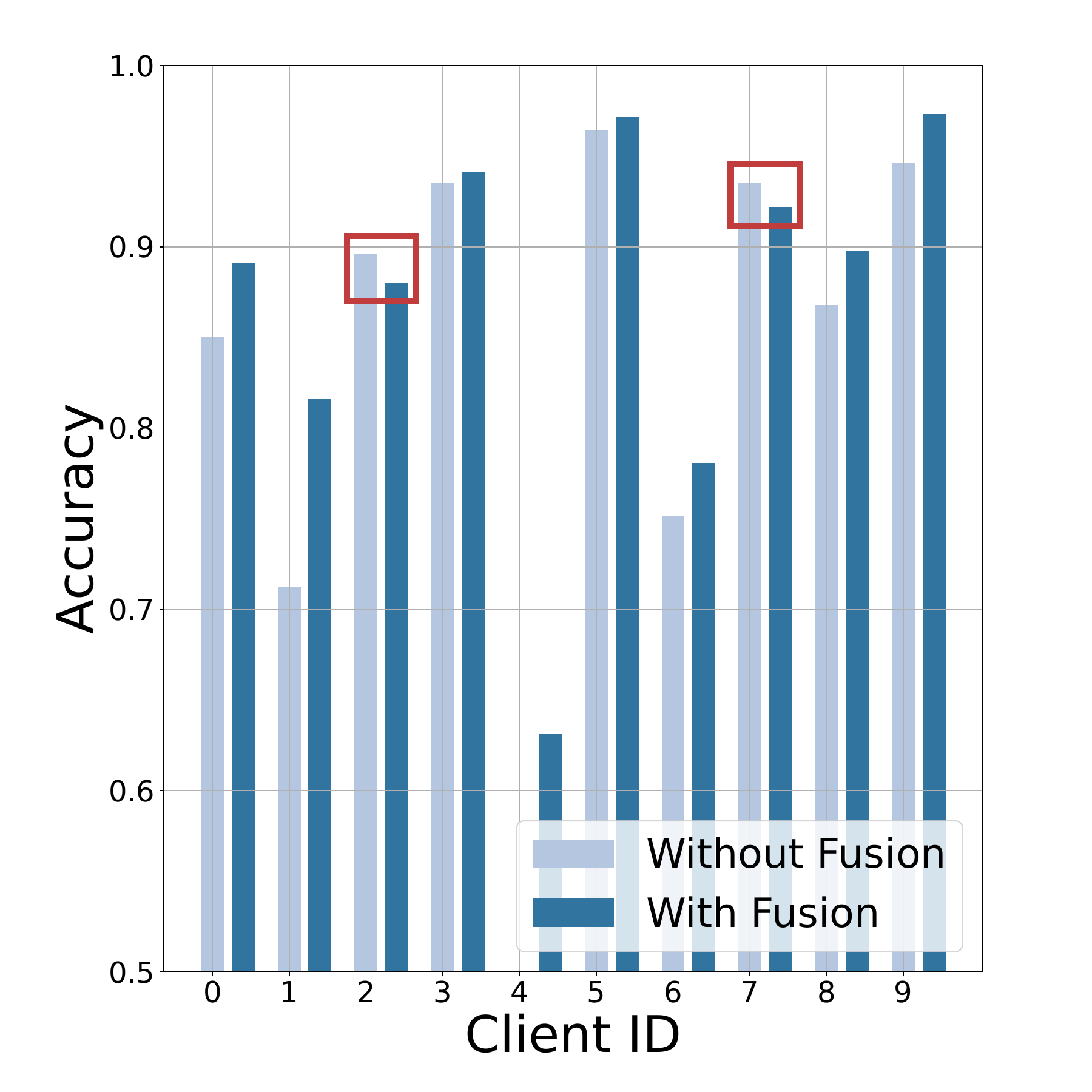}
          \caption{Logits Fusion Client-wise}
          \label{fig3b}
      \end{subfigure}    
    \caption{{Investigation on Personalization Effect in Federated Models.} We visualize the performance improvement after model personalization for both G-FL and P-FL methods. To further investigate such an improvement, we count the number of decision changes and classify them into correction (i.e., from false to true judgment) and mischange (i.e., true to false). Surprisingly, though improvement has been achieved, personalization still introduces significant misjudgment into local models, failing to integrate generic and personalized information for inference effectively. Such a gap severely hampers the federated model to achieve more precise decisions.}
    \label{investigation_gpfl}
\end{figure}

\section{Exploration Experiments}
The exploration experiments are motivated by such a question: \textbf{As the performance of personalized federated learning significantly surpasses that in generic federated learning, have generic models lost their effectiveness? Is it meaningful to integrate generic information? }

Considering this question, we systematically investigate the performance gap between generic and personalized federated models. The exploration experiments were conducted on popular federated benchmarks Cifar10 and Cifar100, and Non-IID data was split among 10 clients. We adopt two strategies to get well-trained personalized federated models. In the first strategy, we utilize a popular personalized federated framework (e.g., FedRoD) to get well-performed personalized models directly. However, in the second strategy, a commonly used model-split strategy has been adopted for global models of G-FL (e.g., FedAvg) to get personalized models.

We report the key observation in Fig. \ref{investigation_gpfl}. Our observations are: (1) Many 'p-hard' instances can be easily classified correctly by generic models but misclassified by personalized models. (2) Though mischanges have been made, personalized models always obtain significant performance improvement as more corrections have been conducted. These observations indicate that the generic model has not lost its effectiveness, calling for the effective integration of personalized and generic information in inference to improve the model's performance further.

However, the fusion of personalized and generic information can be challenging due to the label distribution skew occurring in heterogeneous data. We validate our assertion through validation experiments of popular fusion-based federated methods. Specifically, we conduct both logit fusion and feature fusion, which are introduced in FedRoD \cite{FedRoD} and FedFusion \cite{FedFusion} respectively, and examine the fusion impact of these techniques. Results are reported in Fig. \ref{fig3a} and \ref{fig3b}. As demonstrated, logits fusion and feature fusion inevitably introduce harmful impacts on several clients, causing performance degradation in several clients.
Moreover, diving into the class-wise, such a degradation consistently occurs in the high-accuracy class, which is always the majority class with a large proportion of instances. Such an observation can be interpreted through the heterogeneity impact of federated frameworks. Generic information produced by global models is always more comprehensive than personalized information made by local models. Since the catastrophic forgetting occurred in the minority classes, personalized information is always knowledgeable and professional in its majority classes, while it remains ignorance in the minority classes. So, after fusion, the weaker one will improve. However, as a result, its knowledgeable judgments on its majority classes will also be affected by the introduced noise. \textbf{Thus, how to effectively integrate the generic and personalized information to utilize the comprehensive nature of global models without introducing noise is the key to improving the performance of federated models.} That motivates us to apply uncertainty-based integration to fuse generic and personalized information.

\section{Discussions}
\subsection{Deep Understanding of Instance Uncertainty-Informed Fusion}

Instance uncertainty-informed opinion fusion can be viewed as a weighted average on evidence and prior, treating the uncertainty as weights.

\begin{proposition}
    The evidence of aggregated opinion calculated through instance uncertainty-informed opinion fusion is equivalent to the confidence-based weighted average on observed evidence, that is:
    \begin{equation}
        e_Y^{A \widehat{\diamond} B} = \frac{\boldsymbol{e}_Y^A\left(1-u_Y^A\right)+\boldsymbol{e}_Y^B\left(1-u_Y^B\right)}{2-u_Y^A-u_Y^B}
    \end{equation}

\end{proposition}

\subsection{Discussion about RIPFL}
RIPFL\citep{RIPFL} is a similar work to ours. However, the differences between RIPFL and TPFL are distinct:
\begin{itemize}
    \item TPFL utilizes subjective logic to form opinions, while RIPFL utilizes evidence theory, which can be viewed as a simplified version of subjective logic with union prior. However, in Non-IID data, it is reasonable and necessary to incorporate prior knowledge into the opinion inference. To the best of our knowledge, we are the first to establish a training framework of subjective models, especially in Non-IID data, which could easily lead evidential and subjective models to gradient explosion. (Eq.\ref{evi_reg} provides training stability in Non-IID data.) 
    \item TPFL establishes the concept of model uncertainty and ensures training safety, while RIPFL does not. RIPFL focuses on interpretability. 
    \item TPFL cooperates opinions via instance-uncertainty guided fusion, while RIPFL directly conducts evidence accumulation.
    \item TPFL validates its training and effectiveness in practical severe Non-IID scenarios, while RIPFL mainly focuses on IID data and mitigatory pathological heterogeneous data.

\end{itemize}

\section{Additional Experiments}
\subsection{Experimental Setup}
\label{ExperimentalDetails}
\textbf{Dataset.} We conduct experiments on various popular federated benchmarks, including Cifar10, Cifar100, and Tiny-ImageNet. We utilize latent Dirichlet distribution $\operatorname{Dir}(\beta)$ to partition original training sets to multiple clients to simulate the practical Non-IID scenarios. We choose SVHN as OOD datasets for Cifar10/100. For domain-shift scenarios, We utilize the corruption datasets Cifar10-C and Cifar10-P to validate the reliability of TPFL. For Cifar10, we train a four-layer CNN and further utilize ResNet-18 on Cifar100 and Tiny-ImageNet datasets.

\textbf{Compared Methods.} Due to the trustworthy property of TPFL, we design the comparison experiments from three different aspects. For performance, we compare TPFL with various SOTA generic and personalized federated methods, including FedAvg \cite{FedAvg}, FedProx \cite{FedProx}, MOON \cite{MOON}, FedNTD \cite{FedNTD}, Per-FedAvg \cite{Per-FedAvg}, FedRep \cite{FedRep}, pFedMe \cite{pFedMe}, Ditto \cite{ditto}, FedRoD \cite{FedRoD}, FedFomo \cite{FedFOMO}, APPLE \cite{APPLE}, FedALA \cite{FedALA}. RIPFL \cite{RIPFL} can not converge in Non-IID scenarios, so the comparison between RIPFL is absent. For reliability, we compare the reliability of TPFL against traditional federated frameworks cooperated with uncertainty estimation methods, including popular uncertainty estimation methods MC-Dropout \cite{Dropout}, Deep Ensembles \cite{DeepEnsemble}, and recently-proposed federated framework APH \cite{APH}. For security, we test the robustness of TPFL against prevalent malicious attacks. 
We consider the following threat models under different settings to introduce more sophisticated attacking scenarios.
(1) white-box, full knowledge and online attacks (Type I);
(2) white box, partial knowledge, and online attacks (Type II);
(3) black-box, partial knowledge, and offline attacks (Type III).
Attackers know the model architecture in a white-box setting, while the black box does not. Full knowledge allows attackers to know benign updates and distribution, whereas partial knowledge does not. Online attacks are launched repeatedly during training, while offline attacks occur only at the start. For malicious attacks, we utilize Flip \cite{FlipAttack} (Type III), Random \cite{Random} (Type II), LIE \cite{LIE} (Type I), MPAF \cite{MPAF} (Type II), STAT-OPT \cite{LocalModelPoisoning} (Type I) to launch attacks. For defense strategies, we compare our TPFL with various SOTA defense methods, including Median \cite{MedianMeanTrMedian}, Trimmed-Mean \cite{MedianMeanTrMedian}, Krum \cite{Krum} and Multi-Krum \cite{Krum}.

\textbf{Hyper-parameters.} All experiments are conducted under the following default hyper-parameters without explicit declaration. Following previous work \citep{FedNTD, FedALA, FedRoD, FedKA}, we set the client number to 10 and set $\beta=0.1$ to simulate the critical heterogeneous data scenarios. To ensure the convergence of federated methods and balance the communication cost, we set the local epoch number to 10 with communication round to 100 and 20 for Cifar10/100 and Tiny-ImageNet, respectively. We utilize SGD optimizer to train local models with full participation and further set the learning rate to 0.01. As illustrated in their papers, we tune and choose the best results from the recommended settings for the hyper-parameters involved in comparison experiments. 

\textbf{Balancing Strategy} We utilize a balancing strategy to get balanced local data. Concretely, we set the filter number first. For the down-balanced dataset, we choose the minimal sample number of each class as the threshold among all numbers that are bigger than the threshold and then conduct random samples for those classes whose sample number is larger than the threshold. Other classes are directly selected without replacement. For the up-balanced dataset, except for those classes whose sample number is lower than the threshold, the other classes are sampled to the biggest number with replacement. We choose the best filter number between 20, 50, and 100.

\subsection{Convergence of TPFL}
\begin{figure}
    \centering
    \begin{subfigure}{0.22\textwidth}
    \includegraphics[width=\textwidth]{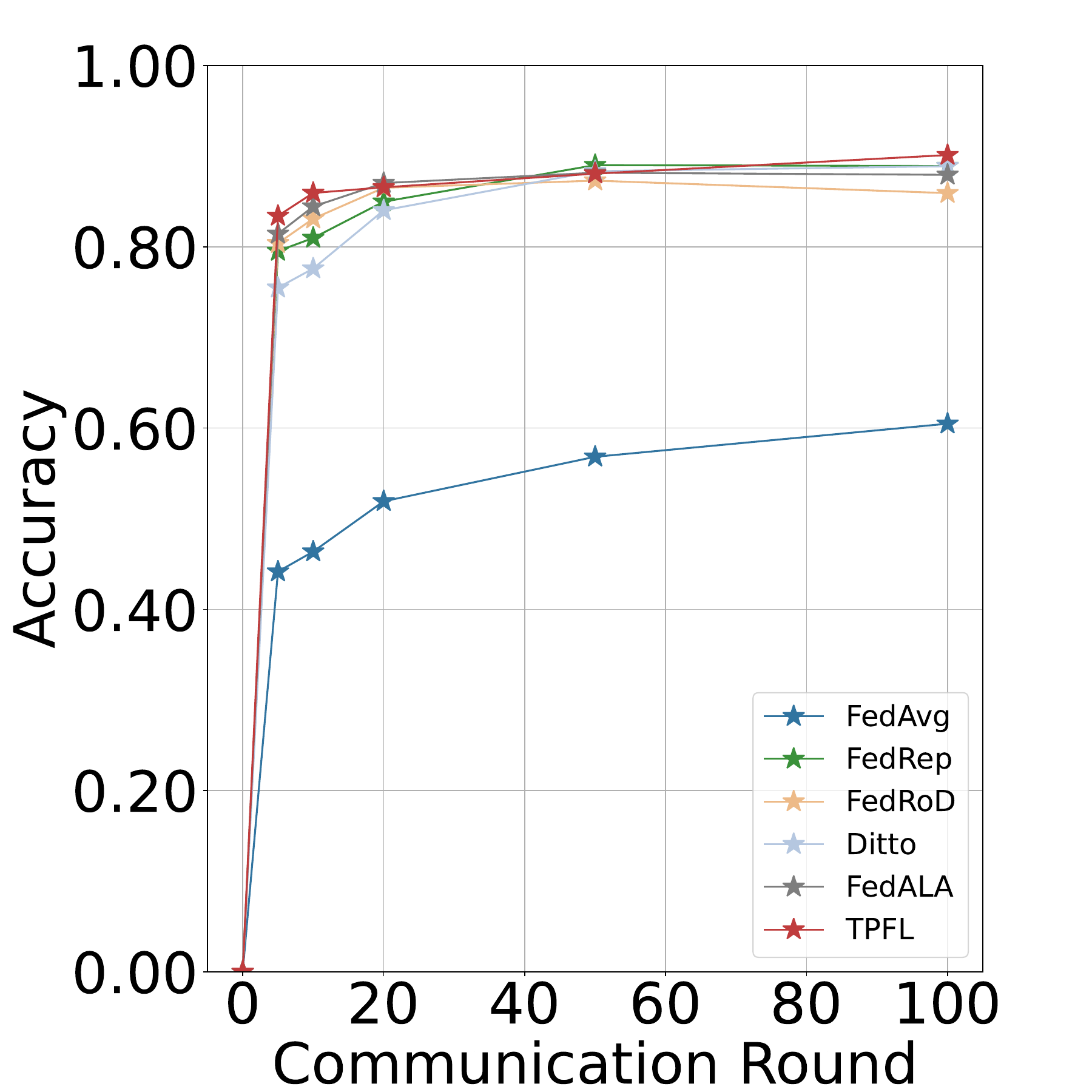}
    \caption{Cifar10 ($\beta=0.1$)}
    \end{subfigure}
    \begin{subfigure}{0.22\textwidth}
    \includegraphics[width=\textwidth]{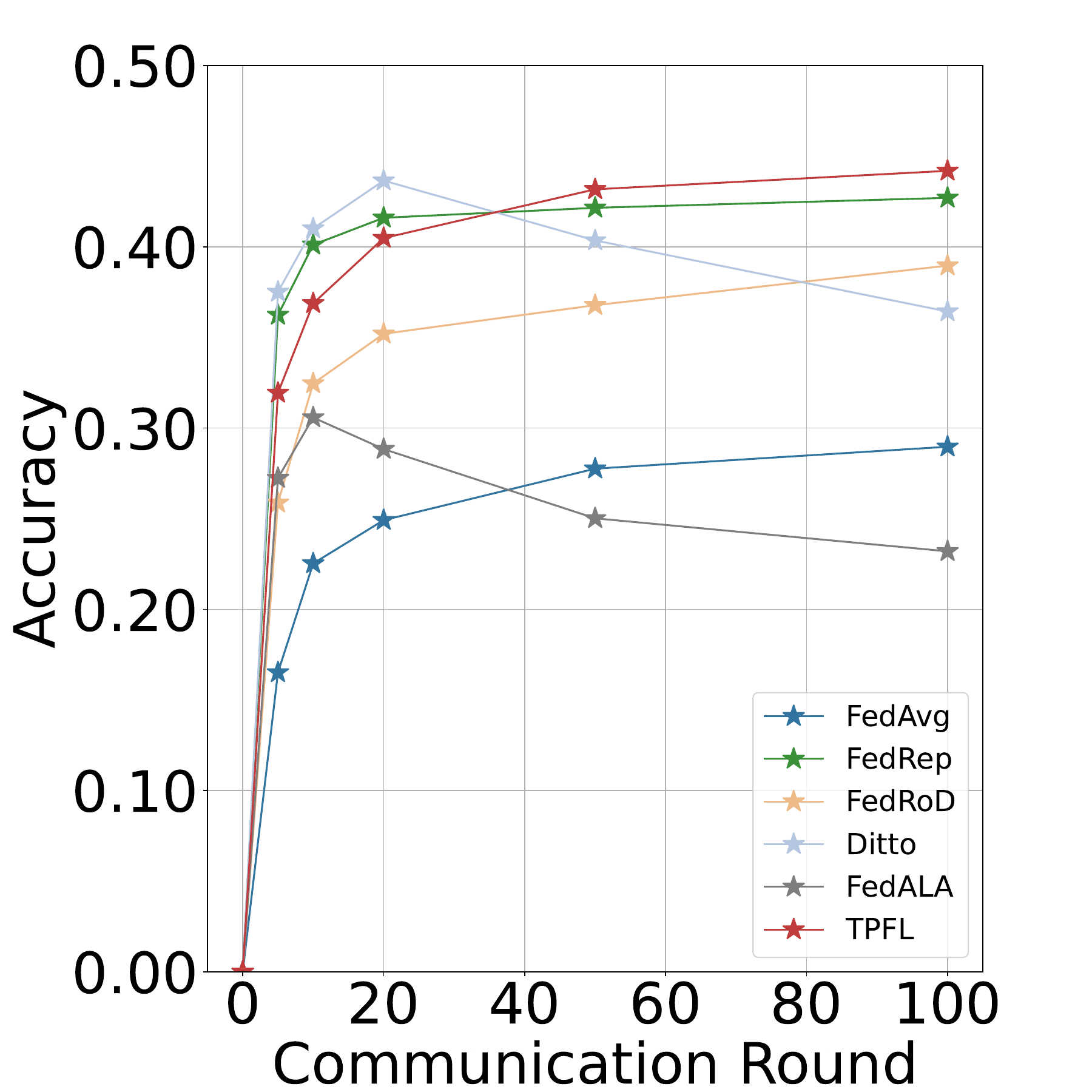}
    \caption{Cifar100 ($\beta=0.1$)}
    \end{subfigure}
    \caption{Convergence curve of TPFL compared with other federated methods in heterogeneous Cifar10/100. As can be seen, TPFL achieves fast and stable convergence among various baselines.}
    \label{convergence}
\end{figure}

We conduct experiments in heterogeneous Cifar10 and Cifar100, further reporting the convergence curve of TPFL in Fig.\ref{convergence}. As can be seen, TPFL achieves fast and stable convergence.

\subsection{Communication cost of TPFL}
TPFL does not involve any additional communication costs. To illustrate this, we choose a proper accuracy threshold as the convergence indicator and report the total communication bandwidth of different federated methods when reaching convergence in various client numbers. Specifically, we set the convergence threshold as 0.85, 0.85, and 0.8 for 10, 20, and 100 clients, respectively. Experimental results are displayed in Table.\ref{communication_tab}.

\begin{table}[htbp]
  \centering
  \caption{Communication cost in Million-Bytes until reaching the desired predictive accuracy under different client sizes.}
  \resizebox{0.7\linewidth}{!}{
    \begin{tabular}{cccc}
    \toprule
    Client Number & Client 10 & Client 20 & Client 100 \\
    \midrule
    FedRep & 1466.37 & 4132.19 & 38652.6 \\
    FedRoD & 1575.14 & 2345.94 & 18767.59 \\
    Ditto & 1675.67 & 3552.43 & 35859.51 \\
    FedALA & 1374.05 & 3150.27 & 32753.48 \\
    TPFL  & 970.13 & 936.8 & 6690.97 \\
    \bottomrule
    \end{tabular}%
    }
  \label{communication_tab}%
\end{table}%

\subsection{Detailed Analysis of TPFL's Computational Cost}

We analyze the computational complexity of each client to evaluate the computational efficiency between TPFL and other FL methods. Conventionally, most FL methods utilize cross-entropy to train the model in the client. For a $K$-classification problem with $n$ samples, the computational complexity of the cross-entropy loss function of each training iteration is $O(Kn)$. For TPFL, its loss function is given as: $ \mathcal{L} = \mathcal{L}_{\text{CE}} + 
 \mathcal{L}_{\text{cor}}
    +
    \lambda_1 \mathcal{L}_{\text{inc}}  +
    \lambda_2
    \mathcal{L}_{\text{evi}} +
    \lambda_3
    \mathcal{L}_{\text{neg}}$. We analyze the complexity of each term one by one:
    \begin{itemize}
        \item $\mathcal{L}_{\text{CE}}$: $O(Kn)$.
        \item $\mathcal{L}_{\text {cor }}(\boldsymbol{x}, \boldsymbol{y})=-u \ln \left(\alpha_{g t}-a_{g t}\right)$: $O(n)$.
        \item $\mathcal{L}_{\text {evi }}=\|\max (\mathbf{0}, \boldsymbol{e}-\epsilon)\|_2^2$: $O(Kn)$, which mainly comes from the searching of the max values.
        \item $\mathcal{L}_{\mathrm{neg}}=\|\boldsymbol{a}-\max (\mathbf{0}, \boldsymbol{a})\|_1$: $O(Kn)$, with an operation of searching max values.
        \item $\mathcal{L}_{\text {inc }}=\ln \frac{\Gamma\left(\sum_{i=1}^k \hat{\alpha}_i\right)}{\Gamma\left(\sum_{i=1}^k a_i\right)}+\sum_{i=1}^k \ln \frac{\Gamma\left(a_i\right)}{\Gamma\left(\hat{\alpha}_i\right)}+ \\
        \sum_{i=1}^k\left(\hat{\alpha}_i-a_i\right)\left[\psi\left(\hat{\alpha}_i\right)-\psi\left(\sum_{i=1}^k \hat{\alpha}_i\right)\right]$: $O(Kn)$. The complexity mainly comes from the calculation of gamma and digamma functions. However, the complexity of gamma and digamma function is $O(1)$ for a single element using numerical approximation, leading to the overall 
 complexity $O(Kn)$ for this loss term.
    \end{itemize}
So the overall complexity is $O(Kn)$.

\onecolumn
\section{Details of Loss Functions}

For Eq.\ref{KL}, we here show its detailed derivation procedure. Given two Dirichlet distribution $\operatorname{Dir}\left(\alpha_1\right)$ and $\operatorname{Dir}\left(\alpha_2\right)$, its KL divergence is: 
\begin{equation}
\begin{aligned}
\mathrm{KL}[P \| Q] & =\left\langle\ln \frac{\frac{\Gamma\left(\sum_{i=1}^k \alpha_{1 i}\right)}{\prod_{i=1}^k \Gamma\left(\alpha_{1 i}\right)} \prod_{i=1}^k x_i{ }^{\alpha_{1 i}-1}}{\frac{\Gamma\left(\sum_{i=1}^k \alpha_{2 i}\right)}{\prod_{i=1}^k \Gamma\left(\alpha_{2 i}\right)} \prod_{i=1}^k x_i{ }^{\alpha_{2 i}-1}}\right\rangle \\
& \left.=\left\langle\ln \left(\frac{\Gamma\left(\sum_{i=1}^k \alpha_{1 i}\right)}{\Gamma\left(\sum_{i=1}^k \alpha_{2 i}\right)} \cdot \frac{\prod_{i=1}^k \Gamma\left(\alpha_{2 i}\right)}{\prod_{i=1}^k \Gamma\left(\alpha_{1 i}\right)} \cdot \prod_{i=1}^k x_i^{\alpha_{1 i}-\alpha_{2 i}}\right)\right\rangle\right\rangle_{p(x)} \\
& =\left\langle\ln \frac{\Gamma\left(\sum_{i=1}^k \alpha_{1 i}\right)}{\Gamma\left(\sum_{i=1}^k \alpha_{2 i}\right)}+\sum_{i=1}^k \ln \frac{\Gamma\left(\alpha_{2 i}\right)}{\Gamma\left(\alpha_{1 i}\right)}+\sum_{i=1}^k\left(\alpha_{1 i}-\alpha_{2 i}\right) \cdot \ln \left(x_i\right)\right\rangle_{p(x)} \\
& =\ln \frac{\Gamma\left(\sum_{i=1}^k \alpha_{1 i}\right)}{\Gamma\left(\sum_{i=1}^k \alpha_{2 i}\right)}+\sum_{i=1}^k \ln \frac{\Gamma\left(\alpha_{2 i}\right)}{\Gamma\left(\alpha_{1 i}\right)}+\sum_{i=1}^k\left(\alpha_{1 i}-\alpha_{2 i}\right) \cdot\left\langle\ln x_i\right\rangle_{p(x)} .
\end{aligned}
\label{KLProof}
\end{equation}
    \label{fig:enter-label}
For a logarithmized Dirichlet variate $x_i$, its expectation is given as:
\begin{equation}
\left\langle\ln x_i\right\rangle=\psi\left(\alpha_i\right)-\psi\left(\sum_{i=1}^k \alpha_i\right)
\end{equation}

Then the Eq.\ref{KLProof} becomes:
\begin{equation}
\mathrm{KL}[P \| Q]=\ln \frac{\Gamma\left(\sum_{i=1}^k \alpha_{1 i}\right)}{\Gamma\left(\sum_{i=1}^k \alpha_{2 i}\right)}+\sum_{i=1}^k \ln \frac{\Gamma\left(\alpha_{2 i}\right)}{\Gamma\left(\alpha_{1 i}\right)}+\sum_{i=1}^k\left(\alpha_{1 i}-\alpha_{2 i}\right) \cdot\left[\psi\left(\alpha_{1 i}\right)-\psi\left(\sum_{i=1}^k \alpha_{1 i}\right)\right]
\end{equation}


\end{document}